\documentclass[runningheads]{llncs}

\usepackage[year=2026]{eccv}

\usepackage{eccvabbrv}

\usepackage{graphicx}
\usepackage{booktabs}
\usepackage{multirow}
\usepackage{amsmath}
\usepackage{amssymb}
\usepackage{overpic}
\usepackage{enumitem}
\usepackage{caption, subcaption}
\usepackage{algorithm}
\usepackage{algorithmic}
\usepackage{wrapfig}
\usepackage{makecell}
\usepackage{ulem}
\usepackage{pgfplots}
\pgfplotsset{compat=1.18}
\usepgfplotslibrary{colorbrewer}

\usepackage[accsupp]{axessibility}  

\usepackage{hyperref}
\hypersetup{colorlinks=false,hidelinks}

\usepackage{orcidlink}

\graphicspath{{./images/}}

\usepackage{float}

\definecolor{Highlight}{HTML}{39b54a} 

\newcommand{\tablestyle}[2]{\setlength{\tabcolsep}{#1}
                            \renewcommand{\arraystretch}{#2}
                            \centering
                            \footnotesize}

\definecolor{plotblue}{RGB}{31,119,180}
\definecolor{plotorange}{RGB}{255,127,14}
\definecolor{plotgreen}{RGB}{44,160,44}
\definecolor{plotred}{RGB}{214,39,40}
\definecolor{plotpurple}{RGB}{148,103,189}

\begin{document}

\title{SuperVoxelGPT: Adaptive and Ordered 3D Tokenization for Autoregressive Shape Generation}

\titlerunning{SuperVoxelGPT}

\author{Yuan Li\inst{1,2} \and
Congyi Zhang\inst{1}$^{\dagger}$ \and
Xifeng Gao\inst{2}$^{\dagger}$ \and
Xiaohu Guo\inst{1}$^{\dagger}$}

\authorrunning{Y.~Li et al.}

\institute{University of Texas at Dallas, USA \and
LightSpeed, USA\\
\email{\{Li.Yuan, congyi.zhang, xguo\}@utdallas.edu, johnyuanli@global.tencent.com, gxf.xisha@gmail.com}\\
$^{\dagger}$~Corresponding authors.}

\maketitle

\begin{center}
  \includegraphics[width=\linewidth]{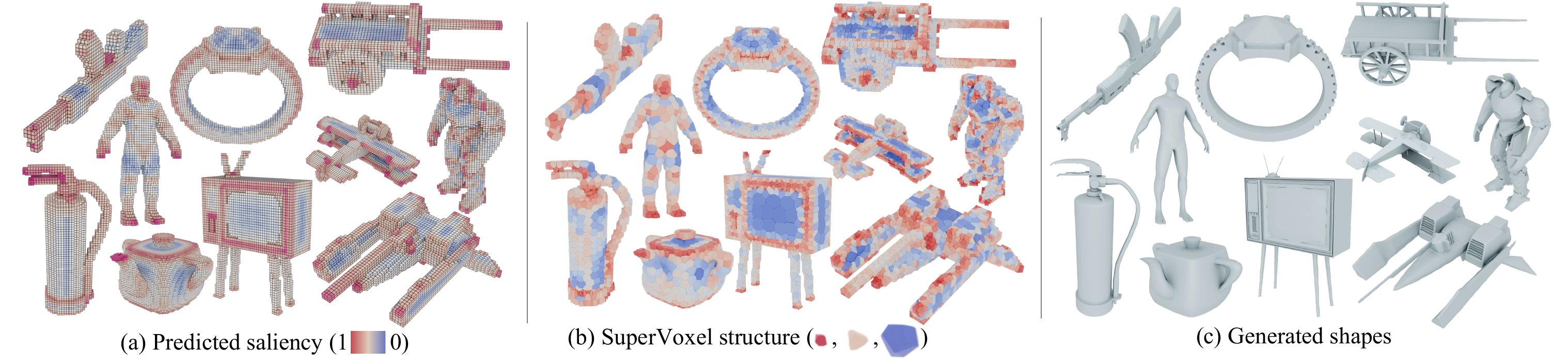}
  \captionof{figure}{We propose a new two-stage MLLM framework for high-resolution 3D generation. The model first predicts a saliency map (a), which is then transformed into our SuperVoxel structure via a customized 3D CVT (b), enabling the extraction of compact and ordered tokens. Finally, our autoregressive model, SuperVoxelGPT, generates high-resolution 3D geometry based on this representation (c).}
  \label{fig:teaser}
\end{center}
\vspace{-9mm}

\begin{abstract}

Autoregressive multimodal large language models (MLLMs) enable 3D generation but struggle to scale to high-resolution shapes due to inadequate 3D tokenizations. Compact set-based representations discard deterministic spatial ordering, leading to ambiguous sequence prediction, while uniform or octree-based voxel grids preserve ordering at the cost of severe redundancy and excessively long sequences. This structural trade-off limits stable and efficient autoregressive 3D generation.
We present \textbf{SuperVoxelGPT}, a representation-first framework that resolves this tension through \emph{adaptive and deterministically ordered supervoxel tokenization}. Given a prompt, we first predict a coarse geometric saliency distribution and construct a shape-adaptive supervoxel partition using saliency-guided centroidal Voronoi tessellation, allocating fine-grained cells to complex regions and larger cells to smooth regions. Conditioned on this prompt and ordered supervoxel layout, we introduce a SuperVoxelVAE and fine-tune a pretrained MLLM to autoregressively generate supervoxel tokens.
Experiments using Trellis-500K data show that SuperVoxelGPT reduces token sequence length to \textbf{12.8\%} of uniform voxel tokenization while achieving state-of-the-art generation quality and an average \textbf{10$\times$} speedup over prior methods.
\end{abstract}

\section{Introduction} \label{sec:intro} 
\begin{figure*}[t]
    \centering
    \begin{overpic}[width=1.0\linewidth]{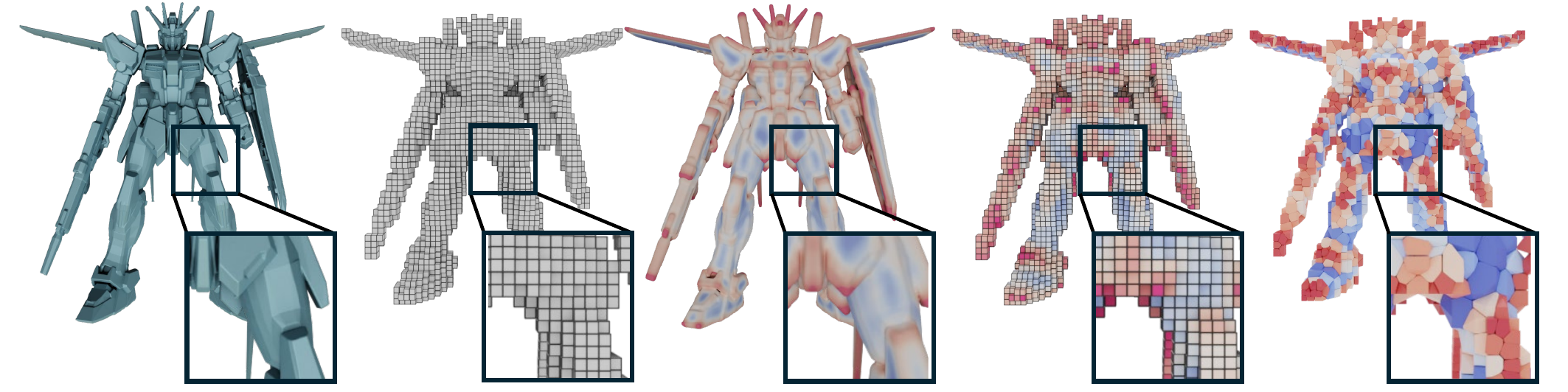}
        \put(8,-2){\textbf{(a)}}
        \put(28,-2){\textbf{(b)}}
        \put(48,-2){\textbf{(c)}}
        \put(68,-2){\textbf{(d)}}
        \put(88,-2){\textbf{(e)}}
    \end{overpic}
    \caption{(a) Original mesh, (b) Uniform voxel structure, (c) Mesh saliency, (d) Volume saliency (e) Supervoxel structure}
    \label{fig:concept}
   \vspace{-3mm}
\end{figure*}
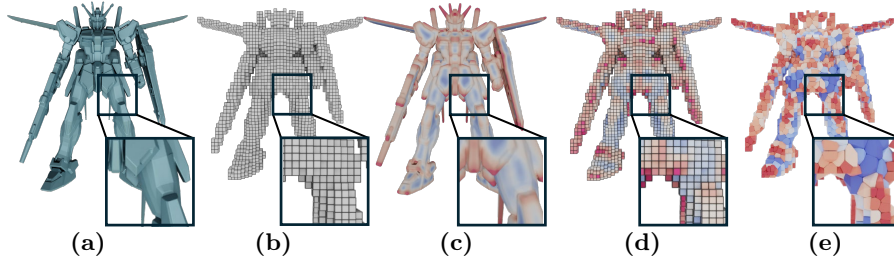
In recent years, Multimodal Large Language Models (MLLMs)~\cite{yin2024survey} have achieved remarkable success across a wide range of generative tasks~\cite{caffagni2024revolution}. Built upon pretrained large language models and fine-tuned on multimodal data, modern MLLMs can jointly reason over text and images, advancing text-to-video~\cite{wang2024gpt4video}, image-to-video~\cite{liu2025dynamic}, and text-to-action generation~\cite{zhang2025openhoi}, exhibiting both state-of-the-art generation quality and speed. In contrast to diffusion-based methods that typically require many iterative denoising steps at training and testing time, autoregressive MLLMs, can generate outputs in several left-to-right passes, often enabling faster inference while maintaining competitive quality~\cite{santilli2023accelerating,tian2024visual}.

Despite these advances, extending MLLMs to high-resolution 3D generation remains challenging. A central reason is that
existing 3D tokenizations are poorly matched to autoregressive modeling, typically falling into two extremes, either overly
compact~\cite{zhang20233dshape2vecset} or overly redundant~\cite{xiang2024structured,wu2025direct3d}, which makes it difficult to achieve both stable sequence modeling and efficient high-resolution generation.

On one end, \emph{compact set-based} tokenizations~\cite{zhang20233dshape2vecset,chen2025dora,lai2025hunyuan3d} maximize information density per token by discarding explicit spatial ordering.  To achieve such high compactness, these methods typically encode the entire geometric shape into a holistic set of latent tokens simultaneously, rather than assigning each token to a specific, localized spatial region. Consequently, the resulting tokens form an unordered collection that lacks a deterministic sequence. This structure is poorly suited to autoregressive modeling, as the absence of a natural order makes sequence prediction ambiguous and weakens the ability of MLLMs to learn stable long-range dependencies. Furthermore, applying quantization on such highly compressed representations inevitably incurs substantial shape information loss, degrading generation quality.

At the other end, \emph{voxel-based} tokenizations~\cite{xiang2024structured,wu2025direct3d,li2025sparc3d} represent the opposite extreme. By discretizing space into a uniform 3D grid and assigning tokens densely to capture local geometry (Fig~\ref{fig:concept}, b), they can preserve high-resolution details with a natural traversal order. However, real shapes exhibit highly non-uniform geometric complexity (Fig~\ref{fig:concept}, c), and uniform grids inevitably waste a large fraction of tokens on smooth or planar regions. This leads to prohibitively long sequences, which undermines inference efficiency and makes autoregressive training harder to scale, less stable, and more prone to degraded generalization~\cite{hou2025focused,mccoy2023embers}. Recently, several methods~\cite{xiang2025native,jia2025ultrashape} have recognized that such long sequences harm both efficiency and training stability, and mitigate this issue by simplifying shapes to a thin layer of surface-intersecting voxels or introducing an additional super-resolution stage to generate the final detailed geometry. However, such simplifications are still not thorough enough: these methods continue to rely on fixed-sized volume tokens to represent shapes, and thus still suffer from inefficiency and instability of redundant token allocation.

To preserve both the efficiency and generation quality of MLLM-based 3D generation, we argue that a new 3D representation
explicitly designed for autoregressive MLLM-based 3D generation is required. And such a representation should satisfy two properties: (1) a \emph{deterministic
sequential structure} that consistently aligns token order with the shape~\cite{wei2025octgpt}, and (2) \emph{adaptive capacity allocation} that concentrates tokens where geometry is complex while avoiding waste where geometry is simple~\cite{lee2005mesh,song2014mesh}. In this way, the tailored representation can remain compact and sequential, making it suitable for autoregressive prediction.

Motivated by these observations, we propose a new representation based on \emph{SuperVoxels}. SuperVoxels partition
space into variable-sized cells (Fig~\ref{fig:concept}, e), naturally enabling adaptive token allocation: detailed regions can be covered by many small
cells, while smooth regions can be represented by fewer large ones~\cite{levy2010p,du2010advances}. Moreover, since each supervoxel is associated with an explicit spatial location, we can impose a deterministic ordering over supervoxels, yielding an autoregressive-friendly 3D token sequence that is both compact and spatially grounded. Compared with unordered set-based tokenizations, this provides a well-defined generation order; compared with uniform or octree-based voxel grids, it substantially reduces redundancy while preserving high-frequency geometric details.

Building on this representation, we introduce \textbf{SuperVoxelGPT}, a two-stage framework for multimodal 3D generation that decouples \emph{where to allocate
tokens} from \emph{what geometry to generate}.

Our main contributions are summarized as follows:
\begin{itemize}
    \item \textbf{SuperVoxelGPT Framework.} We propose a two-stage MLLM framework for high-resolution 3D generation: (i) a prompt-conditioned adaptive
supervoxel partition to allocate tokens by geometric complexity, and (ii) efficient autoregressive generation of supervoxel tokens with Jacobi decoding~\cite{santilli2023accelerating}.
    \item \textbf{SuperVoxelVAE Representation.} We introduce a supervoxel-based 3D tokenizer that produces a compact, deterministically ordered token sequence
aligned with 3D space, making it well suited to stable and efficient autoregressive modeling. Its plug-and-play design allows it to be integrated into any existing sparse voxel VAE to compress token sequences without modifying backbone architectures.
    \item \textbf{Saliency-driven 3D CVT for SuperVoxels.} We develop a customized 3D CVT variant to generate well-shaped, variable-sized supervoxels for
shape-adaptive tokenization. Experiments show that our approach reduces sequence length to 12.8\% of uniform voxel tokenization, achieves the state-of-the-art performance, and delivers a $10\times$ average speedup.
\end{itemize}

\section{Related Works}

\paragraph{3D Representation and Tokenization.}
In recent learning-based 3D generation, popular 3D tokenization schemes broadly fall into two categories: set-based  tokenizations and voxel-based tokenizations.

Set-based methods encode shapes as a compact set of latent tokens, treating geometry as an unordered collection of descriptors. For instance, Shape-2-VecSet~\cite{zhang20233dshape2vecset} embeds sampled surface points and normals into a vector set using a Transformer and supervises decoding with surrounding signed distance field (SDF) values. Subsequent works such as Dora~\cite{chen2025dora} and Hunyuan3D 2.1~\cite{lai2025hunyuan3d} improve sampling strategies by emphasizing salient regions to better preserve fine details. Recently, EfficientTokenizer~\cite{deng2025efficient} further improves the efficiency of vector-set-based tokenizations by using an octree-based adaptive tokenization. While vector-set tokenizations are compact and avoid token redundancy, they commonly exhibit two limitations. First, they are commonly implemented with full self-attention in prior works, which hampers scaling token count for high-resolution geometry. Second, by modeling tokens as an unordered set, they discard deterministic ordering, which introduces permutation ambiguity and makes them less suitable for autoregressive sequence modeling.

Voxel-based tokenizations lie at the opposite end. These methods preserve explicit spatial structure by discretizing space into voxel grids and assigning features to occupied cells. XCube~\cite{ren2024xcube} replaces the global VecSet in 3DShape2VecSet with voxel-aligned SDF and normal features, improving detail preservation. TRELLIS~\cite{xiang2024structured} and Direct3D-S2~\cite{wu2025direct3d} adopt two-stage strategies to avoid generating empty voxels, focusing computation on shape-intersecting regions. TripoSF~\cite{he2025sparseflex}, Sparc3D~\cite{li2025sparc3d}, and TRELLIS2~\cite{xiang2025native} further scale this paradigm to higher resolutions. However, treating each grid cell as an individual token ignores the highly non-uniform distribution of geometric complexity, wasting many tokens on smooth regions and incurring excessive computation.

In contrast, our SuperVoxel-based representation achieves adaptive compactness while preserving a deterministic spatially grounded ordering, making it better suited to autoregressive MLLM-based generation.

\paragraph{3D MLLMs for 3D Generation.} Recent works have explored MLLM-based 3D generation by converting 3D geometry into discrete tokens and autoregressively predicting them. Existing methods fall into two paradigms.

\textit{Token-by-token} methods, such as MeshGPT~\cite{siddiqui2024meshgpt}, MeshAnythingV2~\cite{chen2025meshanything}, PivotMesh~\cite{song2025mesh}, EdgeRunner~\cite{tang2024edgerunner}, LLaMA-Mesh~\cite{wang2024llamameshunifying3dmesh}, and BrickGPT~\cite{pun2025generating}, iteratively predict the next token conditioned on all preceding tokens. However, 3D sequences often reach tens of thousands of tokens, where strictly sequential decoding erases the efficiency advantage over diffusion methods~\cite{deng2025efficient}. Moreover, the variable and unknown sequence length prevents the application of parallel decoding strategies such as Jacobi decoding~\cite{santilli2023accelerating}.

\textit{Sequence-to-sequence} methods recognize the inefficiency of single-token autoregression and instead adopt sequence-to-sequence autoregression to address the efficiency bottleneck~\cite{tian2024visual}. For example, OctGPT~\cite{wei2025octgpt} and Sar3D~\cite{chen2025sar3d} reformulate 3D generation as autoregression from one resolution to the next. However, such approaches typically require uniform resolution levels and explicit inter-resolution correspondences, forcing uniform token allocation and thus producing overly long sequences, which undermines the efficiency gains that sequence-to-sequence methods are designed to achieve.

Our SuperVoxelGPT employs a two-stage design to solve the efficiency bottleneck. In the first stage, we adopt MaskGIT~\cite{chang2022maskgit,chen2025mar3d} to predict the coarse supervoxel structure efficiently, thereby determining the output sequence length a priori. In the second stage, since the sequence structure is already known, we directly apply Jacobi decoding~\cite{santilli2023accelerating} to parallelize the autoregressive generation of the entire supervoxel token sequence, achieving significant speedup of token-by-token autoregression.

\begin{figure*}[t]
    \centering
    \begin{overpic}[width=1.0\linewidth]{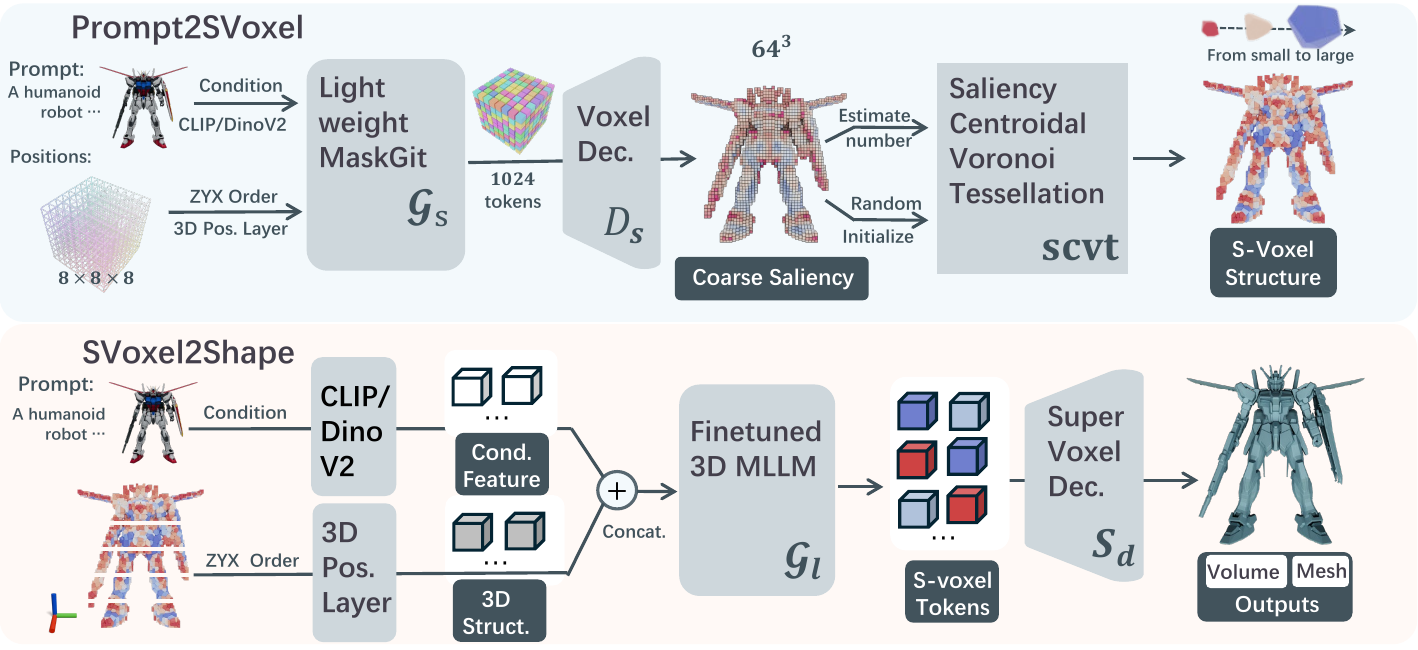}
        \put(1.29, 43.5){\textbf{(a)}}
        \put(1.29, 20){\textbf{(b)}}
    \end{overpic}
        \caption{Overview of SuperVoxelGPT. SuperVoxelGPT employs two stages for 3D asset generation: (a) Prompt-to-supervoxel structure. In the Prompt-to-supervoxel stage, we first predict the coarse saliency distribution of a shape via a lightweight MaskGIT model, then use Saliency-guided Centroidal Voronoi Tessellation to partition the space into adaptive supervoxels based on the saliency distribution. (b) Supervoxel-to-shape stage, we fine-tune an MLLM to generate the supervoxel token sequence guided by the multi-modal prompts and supervoxel positions, and decode the tokens into the final 3D shape. }
    \label{fig:pipeline}
    \vspace{-4mm}
\end{figure*}

\section{Method} \label{sec:method}
As shown in \cref{fig:pipeline}, SuperVoxelGPT consists of two stages: (1) a Prompt-to-supervoxel structure stage and (2) a supervoxel-to-shape generation stage. In the first stage, given a text or image prompt, we predict the geometric complexity distribution of the target shape and partition the 3D space into multiple supervoxels according to the complexity concentration, yielding an adaptive supervoxel structure. In the second stage, we fine-tune an MLLM to autoregressively generate the supervoxel token sequence guided by the multimodal prompt and supervoxel positions, and then decode the tokens into the final 3D shape. We describe the Prompt-to-supervoxel structure stage in \cref{sec:stage1} and the supervoxel-to-shape generation stage in \cref{sec:stage2}.

\subsection{Prompt-to-Supervoxel Structure Stage} \label{sec:stage1}

This stage predicts where to allocate tokens based on the geometric complexity distribution of the target shape. Given a text or image prompt, we first estimate a coarse
saliency distribution indicating regions of varying geometric detail, then partition the 3D space into adaptive supervoxels—allocating more supervoxels to
high-complexity regions and fewer to low-complexity regions. This stage comprises two modules: (1) \emph{Prompt-to-Saliency Volume}, which maps the input prompt to a dense
3D saliency volume encoding the target shape's occupancy and geometric complexity distribution, and (2) \emph{Saliency-guided Centroidal Voronoi Tessellation (SCVT)}, which generates a supervoxel structure adapted to the predicted complexity distribution.

\begin{figure*}[t]
    \centering
    \begin{overpic}[width=1.0\linewidth]{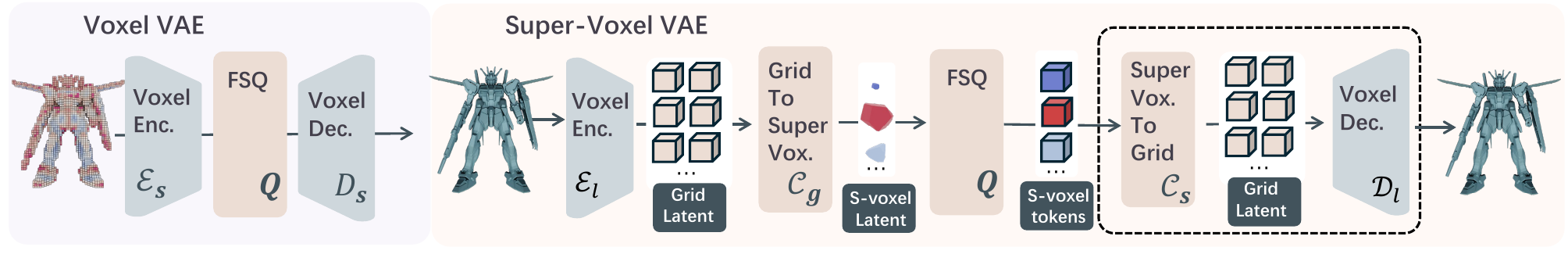}
        \put(1.1, 14){\textbf{(a)}}
        \put(27, 14){\textbf{(b)}}
    \end{overpic}
        \caption{Overview of two VAE structures. (a) Saliency VQ-VAE, (b) SuperVoxelVAE. }
    \label{fig:vaes}
    \vspace{-4mm}
\end{figure*}

\subsubsection{Prompt-to-Saliency Volume}
This module predicts a 3D saliency volume from a text or image prompt using two lightweight components: a Saliency VQ-VAE that encodes dense saliency volumes into 1024 tokens, and a prompt-conditioned MaskGIT model that generates these latent embeddings from the input prompt. We compute ground-truth saliency using
spectral mesh saliency~\cite{song2014mesh} and apply max pooling to produce $64^3$ dense saliency volumes. At training time, we train a 3-layer Saliency VQ-VAE to convert saliency volumes into 1024 tokens, and a MaskGIT~\cite{chang2022maskgit} to generate the token sequence conditioned on the input prompt. At inference time, the MaskGIT generates the token sequence with 12 iterations as suggested in~\cite{chang2022maskgit}, and the decoder reconstructs the saliency volume from the generated tokens.

\textbf{Saliency VQ-VAE.} Our Saliency VQ-VAE inherits the encoder and decoder from TRELLIS2~\cite{xiang2025native} with 3 layers, and adds an FSQ layer in the middle (as shown in~\cref{fig:vaes}(a)). In this way, a $64^3$ saliency volume is encoded into $8 \times 8 \times 8 \times 2 = 1024$ tokens.

Compared to the VAEs in TRELLIS2~\cite{xiang2025native}, our VQ-VAE differs in two aspects: (1) we add a FSQ layer for quantization~\cite{mentzer2023finite}, and (2) we predict not only the occupancy mask but also the saliency values at occupied positions. Prior methods use the last-layer features solely to predict occupancy, while we split the last-layer features into two halves: the first half predicts the occupancy mask, and the second half predicts the saliency values. We inherit the training loss from TRELLIS2~\cite{xiang2025native} and add an additional L1 loss to supervise the saliency values. In this way, a saliency volume is mapped into 1024 tokens.

\textbf{Saliency Volume MaskGIT.} To avoid extensive autoregressive and denoising steps, we adopt the MaskGIT architecture with 24 layers to generate 1024 tokens in only 12 iterations, as suggested in~\cite{chang2022maskgit}. We inherit the original MaskGIT architecture and replace the conditioning inputs with text CLIP features and image DINOv2 features.

\subsubsection{Saliency-guided Centroidal Voronoi Tessellation} \label{sec:cvt}

\begin{wrapfigure}{r}{0.5\linewidth}
    \vspace{-15pt}
    \centering
    \begin{tikzpicture}
    \begin{axis}[
        width=\linewidth,
        height=4.5cm,
        grid=both,
        grid style={line width=0.3pt, draw=gray!25},
        major grid style={line width=0.5pt, draw=gray!40},
        minor tick num=1,
        axis line style={gray!80, line width=0.8pt},
        tick style={gray!80},
        tick label style={font=\small},
        xlabel={$K$ (Maximum SuperVoxel Size)},
        ylabel={Compression Ratio $c$},
        xlabel style={font=\normalsize},
        ylabel style={font=\normalsize},
        title style={font=\bfseries\normalsize, yshift=2pt},
        xmin=0.5, xmax=10.5,
        ymin=0, ymax=1.05,
        legend pos=north east,
        legend style={
            font=\scriptsize,
            draw=gray!50,
            fill=white,
            fill opacity=0.9,
            rounded corners=2pt,
            cells={anchor=west},
            row sep=0pt,
        },
        legend cell align={left},
        every axis plot/.append style={
            line width=0.8pt,
            mark size=1.2pt,
        },
        smooth,
    ]
    \addplot[
        color=plotblue,
        mark=*,
        mark options={solid, fill=plotblue!60!white},
    ] coordinates {
        (1.0,1.000000) (1.5,0.546574) (2.0,0.358059) (2.5,0.257817) (3.0,0.196958) (3.5,0.156783) (4.0,0.128675) (4.5,0.108147)
(5.0,0.092648) (5.5,0.080632) (6.0,0.071111) (6.5,0.063429) (7.0,0.057133) (7.5,0.051904) (8.0,0.047511) (8.5,0.043781)
(9.0,0.040586) (9.5,0.037827) (10.0,0.035426)
    };
    \addlegendentry{$t = 0.1$}
    \addplot[
        color=plotorange,
        mark=square*,
        mark options={solid, fill=plotorange!60!white},
    ] coordinates {
        (1.0,1.000000) (1.5,0.518320) (2.0,0.329743) (2.5,0.233057) (3.0,0.175806) (3.5,0.138708) (4.0,0.113124) (4.5,0.094656)
(5.0,0.080845) (5.5,0.070224) (6.0,0.061867) (6.5,0.055163) (7.0,0.049697) (7.5,0.045178) (8.0,0.041397) (8.5,0.038198)
(9.0,0.035466) (9.5,0.033113) (10.0,0.031071)
    };
    \addlegendentry{$t = 0.2$}
    \addplot[
        color=plotgreen,
        mark=triangle*,
        mark options={solid, fill=plotgreen!60!white},
    ] coordinates {
        (1.0,1.000000) (1.5,0.487837) (2.0,0.299962) (2.5,0.207499) (3.0,0.154291) (3.5,0.120538) (4.0,0.097645) (4.5,0.081337)
(5.0,0.069276) (5.5,0.060084) (6.0,0.052908) (6.5,0.047189) (7.0,0.042553) (7.5,0.038740) (8.0,0.035562) (8.5,0.032885)
(9.0,0.030605) (9.5,0.028648) (10.0,0.026954)
    };
    \addlegendentry{$t = 0.3$}
    \addplot[
        color=plotred,
        mark=diamond*,
        mark options={solid, fill=plotred!60!white},
    ] coordinates {
        (1.0,1.000000) (1.5,0.455905) (2.0,0.269371) (2.5,0.181665) (3.0,0.132832) (3.5,0.102617) (4.0,0.082520) (4.5,0.068428)
(5.0,0.058138) (5.5,0.050380) (6.0,0.044377) (6.5,0.039630) (7.0,0.035808) (7.5,0.032680) (8.0,0.030087) (8.5,0.027911)
(9.0,0.026066) (9.5,0.024486) (10.0,0.023122)
    };
    \addlegendentry{$t = 0.4$}
    \addplot[
        color=plotpurple,
        mark=pentagon*,
        mark options={solid, fill=plotpurple!60!white},
    ] coordinates {
        (1.0,1.000000) (1.5,0.422698) (2.0,0.238298) (2.5,0.155918) (3.0,0.111769) (3.5,0.085247) (4.0,0.068013) (4.5,0.056152)
(5.0,0.047624) (5.5,0.041276) (6.0,0.036416) (6.5,0.032607) (7.0,0.029563) (7.5,0.027089) (8.0,0.025049) (8.5,0.023345)
(9.0,0.021906) (9.5,0.020678) (10.0,0.019621)
    };
    \addlegendentry{$t = 0.5$}
    \end{axis}
    \end{tikzpicture}
    \caption{Compression ratio $c$ as a function of max supervoxel size $K$ for different saliency thresholds $t$, computed from statistics over our training dataset. The monotonic decrease enables efficient parameter selection for target compression ratios.}
    \label{fig:saliency_compress}
    \vspace{-15pt}
\end{wrapfigure}
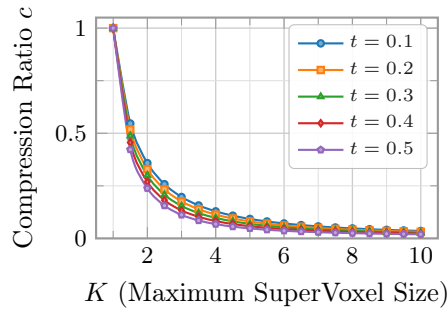
\begin{figure}[t]
    \centering
    \begin{overpic}[width=0.95\linewidth]{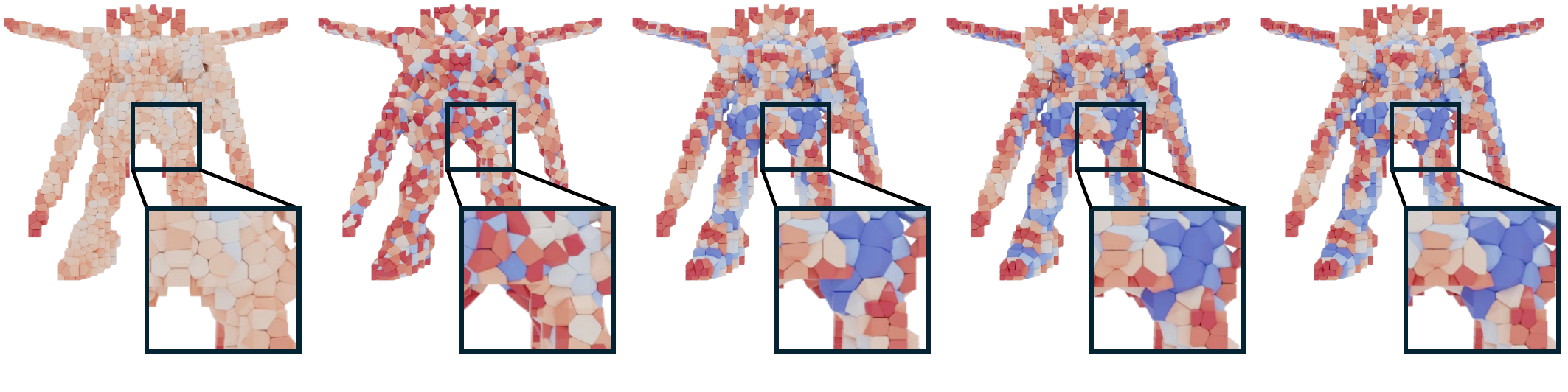}
        \put(9,0){\small (a)}
        \put(29,0){\small (b)}
        \put(49,0){\small (c)}
        \put(69,0){\small (d)}
        \put(89,0){\small (e)}
    \end{overpic}
    \vspace{1mm}
    \caption{Effect of saliency-driven CVT over GPU iterations: (a)~0, (b)~10, (c)~20, (d)~30, and (e)~40 iterations. The supervoxel partition adapts to geometric complexity, with denser cells near detailed features and sparser cells in smooth regions.}
    \label{fig:cvt_effect}
    \vspace{-4mm}
\end{figure}
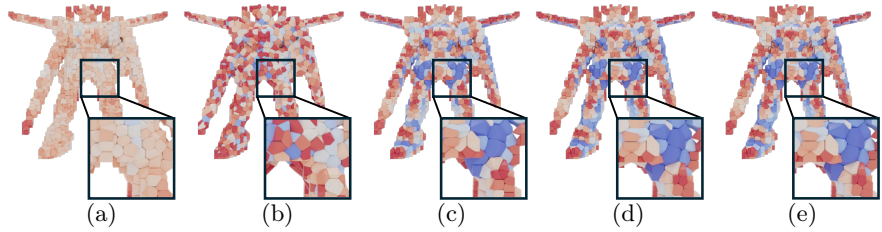
The second module converts the predicted saliency volume into an adaptive supervoxel partition using Centroidal Voronoi Tessellation (CVT)~\cite{du1999centroidal,levy2010p}.
Our goal is to define a piecewise-linear target size field from saliency values to final supervoxel sizes: in geometrically complex regions (high saliency), we maintain unit-size supervoxels with a 1:1 correspondence to grid cells, while in flat or smooth regions (low saliency), we use larger supervoxels of size $K$ to achieve a compact supervoxel structure. 
Since CVT is a well-established algorithm with theoretical convergence guarantees and only requires a density field as input, we carefully design a two-stage mapping from saliency to size field to density field, ensuring that the converged result meets our compression objective.

\paragraph{Saliency-to-Supervoxel Size Field.}
We first design a piecewise-linear mapping from saliency $x$ to target supervoxel size $f(x)$, which determines the desired partition that CVT should converge to. Let $x_i \in [0, 1]$
denote the saliency value at voxel $i$, $t$ is a saliency pivot that separates low-importance regions from salient regions, and $K$ the supervoxel size in low-saliency regions (where $x_i < t$). We
 define the supervoxel size field $f(x; K, t)$ as:
\begin{equation}
    f(x; K, t) = \begin{cases}
    K & \text{if } x < t \\
    \frac{x - t}{1 - t} \cdot (1 - K) + K & \text{if } x \ge t
    \end{cases}
\end{equation}
where $f(x)$ represents the target supervoxel size at saliency $x$. This piecewise linear function ensures that the highest saliency regions ($x=1$) maintain unit
size ($f(1)=1$), while low-saliency regions have supervoxel size $K$. The parameter $K$ controls the compression aggressiveness—larger $K$ means coarser supervoxels
in smooth regions, freeing up more tokens for geometric details.

\paragraph{Supervoxel Size Field to Density Field.}
CVT requires a density field $d(x)$ as input. To make CVT converge to our target size field $f(x)$, we derive the corresponding density: the CVT cell linear size satisfies $f \propto d^{-1/5}$~\cite{du2010advances}, so the density function is:
\begin{equation}
    d(x) \propto \frac{1}{f(x; K, t)^5}
\end{equation}
With this density field, CVT converges to supervoxels matching our target size distribution.

\paragraph{Compression Ratio Control.}
A supervoxel with linear size $f$ occupies approximately $f^3$ voxels, so each voxel contributes $1/f^3$ to the total token count. The estimated number of supervoxels for given parameters $(K, t)$ is:
\begin{equation}
    N(K, t) = \sum_{i \in V} \frac{1}{f(x_i; K, t)^3}
    \label{eq:num_supervoxels}
\end{equation}
where $x_i$ is the saliency value at voxel $i$. We define compression ratio w.r.t. the baseline voxel-tokenization tokens at the same resolution, i.e., $c = N(K,t) / N_{\text{voxels}}$. Since $N(K,t)$ monotonically decreases with $K$ and $t$, we can explore this relationship empirically. 
As shown in \cref{fig:saliency_compress}, increasing $K$ and $t$ progressively coarsens the non-salient regions first and eventually compresses the entire volume, producing a spectrum of compression ratios. Increasing $K$ first yields a rapid decrease in the compression ratio by merging low-saliency regions into larger supervoxels, but the gain gradually saturates once these regions have been largely coarsened; further increasing $K$ then provides limited additional compression and may degrade salient geometry. Thus, $K=4$ offers a good trade-off between compactness and fidelity. Similarly, increasing $t$ suppresses more low-saliency noises, improving compression but risking the removal of genuinely salient structures when set too high. We find $t=0.1$ to be a balanced choice. The parameter sensitivity analysis in the supplementary material further validates this parameter selection. Based on these observations, we select $K=4$ and $t=0.1$ to achieve our target compression ratio of $c=12.8\%$.

\paragraph{CVT Generation Pipeline.}
With the density field $d(x)$ and parameters $(K, t)$, we partition the 3D domain into supervoxels through three steps:
\begin{enumerate}
    \item \textbf{Initialization:} Compute the number of supervoxels $N$ using \cref{eq:num_supervoxels}, and initialize $N$ seeds by uniform sampling from occupied voxels.
    \item \textbf{Lloyd's Iteration:} Iteratively update seed positions using weighted centroids:
    \begin{equation}
        \mathbf{s}_i^{(k+1)} = \frac{\sum_{\mathbf{x} \in V_i} \mathbf{x} \cdot d(\mathbf{x})}{\sum_{\mathbf{x} \in V_i} d(\mathbf{x})}
    \end{equation}
    where $V_i = \{\mathbf{x} : \|\mathbf{x} - \mathbf{s}_i\| \leq \|\mathbf{x} - \mathbf{s}_j\|, \forall j \neq i\}$ is the Voronoi cell of seed $\mathbf{s}_i$, and $d(\mathbf{x})$ is the density at position $\mathbf{x}$. As shown in \cref{fig:cvt_effect}, after 40 iterations the supervoxels concentrate in geometrically complex regions.
    \item \textbf{Ordering:} Sort supervoxel centers by Z-Y-X coordinates to obtain a deterministic sequence $\{\mathbf{p}_1, \mathbf{p}_2, \ldots, \mathbf{p}_N\}$, enabling autoregressive generation.
\end{enumerate}
In this way, we obtain a compact supervoxel structure that is well-shaped and adapted to the geometric complexity distribution of the target shape.

\subsection{Supervoxel-to-Shape Generation} \label{sec:stage2}
Given a text or image prompt and the supervoxel structure from the previous stage, this stage generates the corresponding 3D shapes. It consists of two components: (1) \emph{SuperVoxelVAE}, which learns to encode local geometry into discrete supervoxel tokens, and (2) a fine-tuned \emph{MLLM}, which autoregressively generates the token sequence conditioned on the multimodal prompt. During training, we first train SuperVoxelVAE to learn the shape-to-token mapping, then fine-tune Qwen2.5-0.5B to learn the prompt-to-token mapping. At inference time, the MLLM generates 3D tokens from the input prompt, which are decoded into geometry by the SuperVoxelVAE decoder.

\subsubsection{SuperVoxelVAE}

\begin{figure*}[t]
    \centering
    \includegraphics[width=1\linewidth]{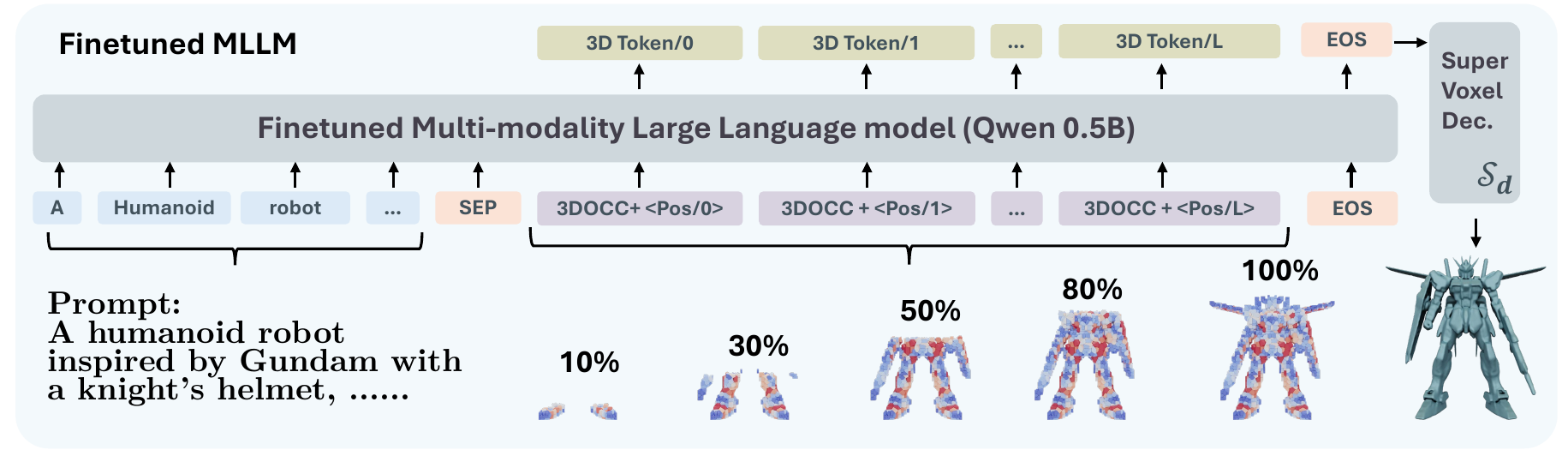}
    \caption{Architecture of the MLLM generation module.  The MLLM autoregressively generates the token sequence conditioned on the multimodal prompt, which is then decoded into the final 3D shape.}
    \label{fig:mllm_structure}
     \vspace{-2mm}
\end{figure*}

SuperVoxelVAE encodes the local 3D geometry within each supervoxel into a compact discrete token. As shown in \cref{fig:pipeline}, it builds upon the sparse grid convolution architecture from prior methods~\cite{xiang2025native,wu2025direct3d} and introduces three additional layers:
\begin{itemize}
    \item \textbf{Grid-to-Supervoxel Layer:} Compresses grid-level latent features into super-voxel level representations.
    \item \textbf{FSQ Layer:} Applies Finite Scalar Quantization~\cite{mentzer2023finite} to convert continuous supervoxel latents into discrete tokens.
    \item \textbf{Supervoxel-to-Grid Layer:} Maps supervoxel latents back to grid structure for decoding.
\end{itemize}
Since these three layers operate solely on the interface between grid latents and supervoxel centers, they are agnostic to the specific sparse convolution backbone. This plug-and-play design allows our supervoxel tokenization to be readily integrated into any existing sparse voxel VAE~\cite{xiang2024structured,wu2025direct3d,xiang2025native}, compressing their token sequences without modifying their encoder-decoder architectures or training losses.

We adopt KNN Cross Attention from PointTransformerV2~\cite{wu2022point} for the grid-supervoxel conversion. Taking the grid-to-supervoxel layer as an example: given grid latents $\mathbf{G}_f$ from sparse convolution, we first initialize supervoxel features $\mathbf{V}_f$ via nearest neighbor search between grid coordinates $\mathbf{G}_c$ and supervoxel centers $\mathbf{V}_c$. Then, KNN Cross Attention refines these features into $\mathbf{V}'_f$ by aggregating information from neighboring grid cells. The decoder uses a symmetric supervoxel-to-grid layer for reconstruction.
Since our architecture is built upon TRELLIS2 with these three plug-in modules, we directly inherit the loss functions and training paradigm from TRELLIS2~\cite{xiang2025native}.

\subsubsection{MLLM-based Token Generation}

With the supervoxel structure $\{\mathbf{p}_1, \mathbf{p}_2, \ldots,\allowbreak \mathbf{p}_N\}$ defined, we formulate 3D generation as an autoregressive next-token prediction task. We fine-tune Qwen2.5-0.5B~\cite{hui2024qwen2} to autoregressively predict supervoxel tokens conditioned on both the multimodal prompt and supervoxel positions. Since the sequence length $N$ is already determined by Stage~1, we can directly apply Jacobi decoding~\cite{santilli2023accelerating} at inference time to parallelize token generation.

The input sequence concatenates prompt embeddings with token embeddings augmented by spatial information:
\begin{equation}
    \mathbf{E}_\text{input} = [\mathbf{E}_\text{prompt}; \mathbf{E}_\text{token} + \text{SA}^2(\text{FourierPE}(\mathbf{p}_1, \ldots, \mathbf{p}_N))]
\end{equation}
where $\mathbf{E}_\text{prompt}$ is the prompt embeddings (from text tokens or image features), $\mathbf{E}_\text{token}$ represents the token embeddings, $\text{FourierPE}(\cdot)$ computes Fourier positional embeddings~\cite{vaswani2017attention} from supervoxel coordinates, and $\text{SA}^2(\cdot)$ denotes a 2-layer self-attention module that processes the positional embeddings before adding them to the token embeddings. This allows the model to capture inter-position relationships and leverage both semantic and spatial information.

\paragraph{Autoregressive Training.}
We train the model with the standard next-token prediction objective. Given the token sequence $(t_1, t_2, \ldots, t_N)$ ordered by the supervoxel structure, the model learns to predict each token conditioned on all preceding tokens, the prompt $\mathcal{T}$, and the supervoxel positions:
\begin{equation}
    \mathcal{L}_\text{MLLM} = -\sum_{i=1}^{N} \log p(t_i | t_{<i}, \mathcal{T}, \mathbf{p}_1, \ldots, \mathbf{p}_N)
\end{equation}

\paragraph{Jacobi Decoding.}
At inference time, standard autoregressive decoding generates tokens one by one, requiring $N$ sequential forward passes. Since our Stage~1 already determines the sequence length $N$, we apply Jacobi decoding~\cite{santilli2023accelerating} to accelerate inference. Jacobi decoding initializes all $N$ token positions with random predictions and iteratively refines them in parallel: at each iteration, every position is updated simultaneously by predicting $t_i$ conditioned on the current estimates of $t_{<i}$. When consecutive tokens converge to a fixed point, they are accepted together, effectively generating multiple tokens per iteration. The decoded tokens are then processed by SuperVoxelVAE decoder to reconstruct the final 3D shape.

\section{Experiments} \label{sec:experiments}

We evaluate SuperVoxelGPT on both text-to-3D and image-to-3D generation tasks. We first describe the experimental setup (\cref{sec:exp_setup}), then compare with state-of-the-art methods (\cref{sec:comparison}), and finally discuss limitations (\cref{sec:limitations}). Ablation studies, parameter sensitivity analysis, discussion, and more quantitative results are provided in the supplementary material.

\subsection{Experimental Setup} \label{sec:exp_setup}

\paragraph{Datasets.}
We select a subset from the Trellis-500K dataset~\cite{xiang2024structured}, using 10,000 shapes for training and 1,000 for testing. We filter out meshes with poor quality by removing those whose non-manifold faces exceed 20\% of the total surface area, as such meshes produce unreliable ground-truth saliency and supervoxel structures.

\paragraph{Evaluation Metrics.}
We evaluate generation quality from four aspects: geometry, semantics, surface detail, and efficiency. For geometric accuracy, we use L2 Chamfer Distance (CD$_{L2}$) and PSNR on rendered multi-view normal images. For semantic alignment, we compute ULIP-2 Similarity, which measures the cosine similarity between generated and ground-truth shapes in a unified 3D-language embedding space. For surface detail quality, we report the area-weighted sum of mean curvature as Mean Curvature Sum (MCS). For efficiency, we report inference time (seconds per shape) and the number of tokens generated. The calculation definitions of these metrics can be found in the supplementary material.

\begin{table}[H]
    \centering
    \caption{Quantitative comparison on text-to-3D generation. $\downarrow$ indicates lower is better, $\uparrow$ indicates higher is better. Best results are in \textbf{bold}, second best are \underline{underlined}. All methods report averages over the test set.}
    \label{tab:text2shape}
    \tablestyle{4pt}{1.1}
    \resizebox{\linewidth}{!}{
    \begin{tabular}{l|ccccccc}
        \toprule
        Method & CD$_{L2}$ $\downarrow$ & PSNR $\uparrow$ & ULIP-2 Sim. $\uparrow$ & MCS $\uparrow$ & Time (s) $\downarrow$ & Resolution & Tokens \\
        \midrule
        BrickGPT~\cite{pun2025generating}        & 0.0851   & 12.99  & 0.199 & \underline{15.39} & 215.5 & $20^3$    & 179 \\
        OctGPT~\cite{wei2025octgpt}              & \underline{0.0797} & \underline{19.80} & \underline{0.329} & 11.74 & \underline{165.8} & $256^3$   & 47,783 \\
        \midrule
        \textbf{Ours}                 & \textbf{0.0134} & \textbf{32.05} & \textbf{0.469} & \textbf{44.19} & \textbf{4.48} & $1024^3$ & \textbf{1,065} \\
        \bottomrule
    \end{tabular}
    }

    \vspace{3mm}

    \caption{Quantitative comparison on image-to-3D generation. Best results are in \textbf{bold}, second best are \underline{underlined}. Resolution and token counts are reported for each method.}
    \label{tab:image2shape}
    \tablestyle{4pt}{1.1}
    \resizebox{\linewidth}{!}{
    \begin{tabular}{l|ccccccc}
        \toprule
        Method & CD$_{L2}$ $\downarrow$ & PSNR $\uparrow$ & ULIP-2 Sim. $\uparrow$ & MCS $\uparrow$ & Time (s) $\downarrow$ & Resolution & Tokens \\
        \midrule
        CraftsMan3D~\cite{li2024craftsman}     & 0.0252 & 22.90 & 0.448 & 37.85 & \underline{7.92}   & $512^3$   & 2,048 \\
        TRELLIS~\cite{xiang2024structured}    & 0.0182 & 27.28 & 0.464 & 21.64 & 9.03   & $256^3$   & 8,147 \\
        Direct3D-S2~\cite{wu2025direct3d}     & 0.0168 & 26.14 & 0.461 & 33.17 & 137.2  & $1024^3$  & 55,420 \\
        TRELLIS2~\cite{xiang2025native}       & \underline{0.0123} & \textbf{32.19} & \textbf{0.475} & \textbf{44.24} & 34.25   & $1024^3$  & 7,303 \\
        \midrule
        \textbf{Ours}              & \textbf{0.0122} & \underline{32.08} & \underline{0.474} & \underline{44.21} & \textbf{4.60} & $1024^3$ & \textbf{1,048} \\
        \bottomrule
    \end{tabular}
    }
\end{table}

\paragraph{Implementation Details.}
Our method consists of two generative modules (Mask\-GIT and fine-tuned MLLM) and two VAEs (Saliency VQ-VAE and SuperVoxelVAE). The Saliency Volume MaskGIT follows the MaskGIT architecture with 24 layers conditioned on text CLIP features and image DINOv2 features, generating tokens in 12 iterations at inference. The MLLM is fine-tuned from Qwen2.5-0.5B~\cite{hui2024qwen2} with the standard next-token prediction objective and applies Jacobi decoding~\cite{santilli2023accelerating} at inference. Since the number of supervoxels is known from Stage~1, we set the Jacobi context window to cover all supervoxel tokens so that all tokens can be updated in parallel, and cap the iterations to 30. The Saliency VQ-VAE uses 3 encoder/decoder layers with an FSQ layer (num\_levels $= [9, 9, 5, 5]$, num\_quantizer $= 2$), encoding each $64^3$ saliency volume into 1024 tokens. The SuperVoxelVAE is built upon TRELLIS2~\cite{xiang2025native} with plug-in grid-to-supervoxel, FSQ, and supervoxel-to-grid layers. We set the KNN neighborhood size for grid-supervoxel conversion to 16 and the FSQ codebook levels to $[9, 9, 5, 5, 5]$.
We discretize the density field on a  $256^3$ grid (upsampled from $64^3$) to approximate the CVT energy and weighted centroids. All trainable modules are trained for 300 epochs with a learning rate of $5 \times 10^{-5}$ and 5,000 warm-up steps on 8$\times$H100 GPUs (AMD EPYC 9334 32-Core Processor). All inference time measurements are conducted on a single NVIDIA RTX 4090.

\paragraph{Baselines.}
For text-to-3D, we compare with BrickGPT~\cite{pun2025generating} and OctGPT~\cite{wei2025octgpt}. For image-to-3D, we compare with CraftsMan3D~\cite{li2024craftsman}, TRELLIS~\cite{xiang2024structured}, Direct3D-S2~\cite{wu2025direct3d}, and TRELLIS2~\cite{xiang2025native}. Detailed parameter settings for all baselines can be found in the supplementary material.

\subsection{Comparison with State-of-the-Art Methods} \label{sec:comparison}

\cref{tab:text2shape} and \cref{tab:image2shape} present quantitative comparisons on the text-to-3D and image-to-3D generation tasks, respectively.

\begin{figure*}[t]
    \centering
    \includegraphics[width=1\linewidth]{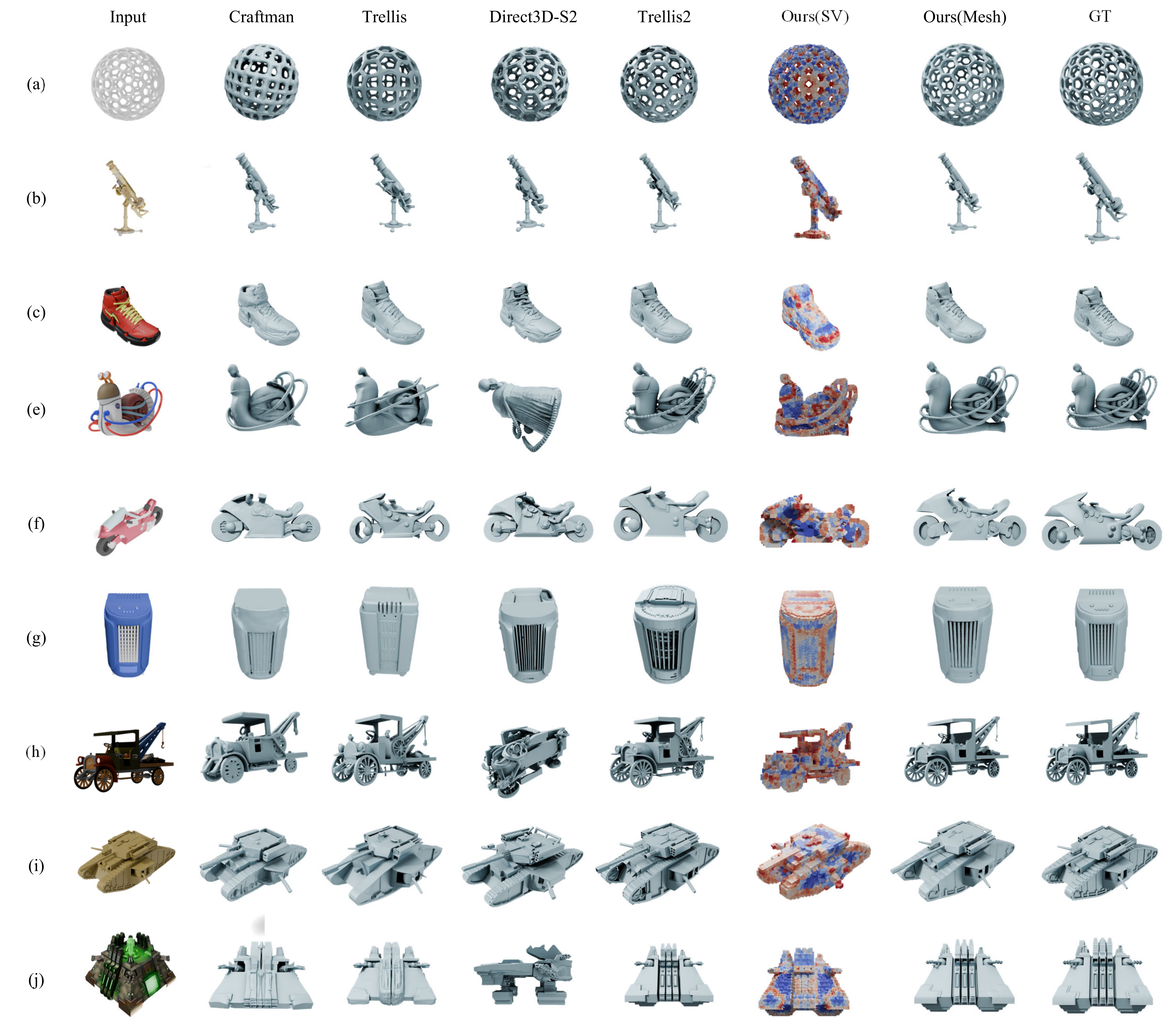}
    \caption{Qualitative comparison on image-to-3D generation. Given an input image, we compare the generated 3D shapes from CraftsMan3D, TRELLIS, Direct3D-S2, TRELLIS2, and our method (SV, i.e., supervoxel structure). }
    \label{fig:image2shape}
    \vspace{-4mm}
\end{figure*}
\paragraph{Generation Quality.}
Our method achieves generation quality on par with the best existing methods across both text-to-3D and image-to-3D tasks in terms of geometric accuracy, semantic alignment, and surface detail. Taking the image-to-3D task as an example (\cref{tab:image2shape}), our method, TRELLIS2, and Direct3D-S2 all operate at $1024^3$ resolution, which enables richer geometric detail and stronger generation performance compared to lower-resolution methods such as CraftsMan3D ($512^3$) and TRELLIS ($256^3$). Moreover, since our SuperVoxelVAE inherits the multi-view normal rendering loss from TRELLIS2~\cite{xiang2025native}, the generated shapes are sharper and more geometrically faithful compared to Direct3D-S2. As shown in \cref{tab:image2shape}, our method closely matches TRELLIS2 across all quality metrics, with only a marginal gap in multi-view normal PSNR ($-$0.11~dB). A minor limitation is that regions with saliency below $t=0.1$ receive a much lower token budget due to aggressive coarsening, which may wash out very fine-grained structures---\eg, the ring-shaped grooves on the pipe in \cref{fig:image2shape}(e) and the spoke grooves on the wheel hub in \cref{fig:image2shape}(f). Lowering $t$ could alleviate this issue but would reduce the compression ratio and make the method more susceptible to saliency-prediction noises. This explains why our method is slightly inferior to TRELLIS2 on particularly fine details.

For the text-to-3D task (\cref{tab:text2shape}), our method also demonstrates strong performance. Compared to BrickGPT and OctGPT, which are limited to low-resolution generation and constrained to specific shape categories (\eg, LEGO structures), our method supports general-purpose shapes and scales to $1024^3$ resolution, achieving substantially better results across all metrics.

\paragraph{Generation Efficiency.}
Our method achieves a significant speedup over existing approaches. Compared to the average inference time of all image-to-3D baselines (47.1s), our method reduces inference time by $10\times$, requiring only 4.60s to generate a $1024^3$-resolution shape. The inference time breaks down as follows: CVT takes 1.209s, the SuperVoxelVAE decoder takes 1.147s, the MLLM takes 1.202s, and the MaskGIT and Saliency VQ-VAE decoder contribute 0.805s and 0.237s, respectively. We attribute this speedup to two factors. First, our token generation is highly parallel: the MaskGIT natively generates all saliency tokens in parallel, and the MLLM leverages Jacobi decoding to generate multiple shape tokens per iteration once the supervoxel count is known. This avoids the sequential bottleneck of generating one token at a time. Second, our supervoxel tokenization produces significantly fewer tokens than competing methods (\eg, 1,048 \vs 55,420 for Direct3D-S2 and 47,783 for OctGPT), directly translating into faster autoregressive generation.

\subsection{Limitations} \label{sec:limitations}

While SuperVoxelGPT achieves strong performance with compact token sequences, several limitations remain:
\begin{itemize}[nosep]
    \item \textbf{Saliency prediction dependency:} The quality of the final shape depends on the accuracy of the first-stage saliency prediction. As illustrated in \cref{fig:image2shape}(c), since saliency values below $t=0.1$ are  heavily coarsened, our method is insensitive to structures with very low geometric saliency, such as the mesh-like perforations on the shoe. Because these regions are not identified as salient, only sparse tokens are allocated to reconstruct them, causing such fine structures to be smoothed over.
    \item \textbf{Degradation on detail-rich shapes:}  For shapes where fine details are uniformly distributed across the surface, the saliency-guided allocation degenerates towards a near-uniform distribution, reducing the compression advantage. 
    \item \textbf{Geometry-only generation:} our supervoxel-based compression is currently designed for geometry and does not extend to texture modeling; thus, our framework cannot generate or represent surface textures or material properties. In addition,  our method is trained on only 10,000 shapes from the Trellis-500K dataset, which may limit its generalization compared to methods trained on larger datasets.

\end{itemize}

\section{Conclusion}

We presented SuperVoxelGPT, a representation-first framework for high-resolution 3D generation that mitigates the fundamental tension between sequence compactness and geometric fidelity in autoregressive 3D generation. Our key insight is to decouple \emph{where to allocate tokens} from \emph{what geometry to generate}. Our experiments demonstrate that SuperVoxelGPT compresses the token sequence to 12.8\% of uniform voxel tokenization while achieving generation quality on par with state-of-the-art methods and a $10\times$ inference speedup, validating that adaptive tokenization is a promising direction for scalable and efficient 3D generation. Moreover, the plug-and-play nature of our supervoxel layers makes them readily applicable to other sparse voxel VAE frameworks, offering a general-purpose compression module for autoregressive 3D generation.

\bibliographystyle{splncs04}
\bibliography{src/ref/reference}

\clearpage
\appendix
\setcounter{section}{0}
\renewcommand{\thesection}{\Alph{section}}
\renewcommand{\thetable}{\Alph{section}\arabic{table}}
\renewcommand{\thefigure}{\Alph{section}\arabic{figure}}
\renewcommand{\theequation}{\Alph{section}\arabic{equation}}
\setcounter{table}{0}
\setcounter{figure}{0}
\setcounter{equation}{0}

\begin{center}
  {\Large\bfseries Supplementary Material}
\end{center}
\vspace{4mm}


\section{Metrics Calculation} \label{app:sec:metrics}

We provide detailed definitions of the evaluation metrics used in the main paper. We first describe the shape alignment procedure (\cref{app:sec:alignment}), which is a prerequisite for most metrics, and then define each metric in detail (\cref{app:sec:metrics_def}).

\subsection{Alignment} \label{app:sec:alignment}

Given a generated shape $\mathcal{A}$ and the corresponding ground-truth shape $\mathcal{B}$, we apply all $2^3 = 8$ axis-flip combinations along the $x$, $y$, and $z$ coordinate axes to $\mathcal{A}$, producing eight candidate orientations $\{\mathcal{A}_k\}_{k=1}^{8}$. For each candidate, we uniformly sample $100{,}000$ points from both $\mathcal{A}_k$ and $\mathcal{B}$, and register $\mathcal{A}_k$ to $\mathcal{B}$ using Fast Global Registration (FGR)~\cite{zhou2016fast} for coarse alignment followed by Iterative Closest Point (ICP) for refinement. We then compute the $L_2$ Chamfer Distance between each aligned pair and retain only the shape $\mathcal{A}^{*}$ that yields the smallest Chamfer Distance. All subsequent shape-comparison metrics are evaluated using $\mathcal{A}^{*}$ and $\mathcal{B}$.

\subsection{Metrics} \label{app:sec:metrics_def}

\paragraph{$L_2$ Chamfer Distance ($\mathrm{CD}_{L_2}$).}
We uniformly sample $100{,}000$ points from the surfaces of the aligned generated mesh $\mathcal{A}^{*}$ and the ground-truth mesh $\mathcal{B}$, yielding point sets $P$ and $Q$, respectively. The $L_2$ Chamfer Distance~\cite{fan2017pointsetgen} is defined as:
\begin{equation}
    \mathrm{CD}_{L_2}(P, Q) = \frac{1}{|P|}\sum_{\mathbf{p} \in P} \min_{\mathbf{q} \in Q} \|\mathbf{p} - \mathbf{q}\|_2^2 \;+\; \frac{1}{|Q|}\sum_{\mathbf{q} \in Q} \min_{\mathbf{p} \in P} \|\mathbf{q} - \mathbf{p}\|_2^2,
\end{equation}
where the two terms measure the average squared nearest-neighbor distance from $P$ to $Q$ and from $Q$ to $P$, respectively. Lower $\mathrm{CD}_{L_2}$ indicates better geometric reconstruction quality.

\paragraph{PSNR on Multi-View Normal Maps.}
We render multi-view normal maps from 24 random viewpoints for both the aligned generated and ground-truth shapes using the PyTorch3D rasterizer~\cite{ravi2020pytorch3d}, with the camera radius set to $2$ and the field of view (FoV) set to $40^\circ$. The Peak Signal-to-Noise Ratio (PSNR) is computed per view and averaged:
\begin{equation}
    \mathrm{PSNR} = \frac{1}{V}\sum_{v=1}^{V} 10 \cdot \log_{10}\!\left(\frac{\mathrm{MAX}^2}{\mathrm{MSE}_v}\right),
\end{equation}
where $V$ is the number of rendered views, $\mathrm{MAX}$ is the maximum possible pixel value, and $\mathrm{MSE}_v$ denotes the mean squared error between the normal-map renderings of view $v$. Higher PSNR indicates better geometric surface fidelity.

\paragraph{ULIP-2 Similarity.}
ULIP-2 Similarity~\cite{xue2024ulip2} measures the semantic alignment between a generated shape and its reference description in a unified vision-language-3D embedding space. We extract ULIP-2 feature embeddings $\mathbf{f}_{\mathrm{gen}}$ and $\mathbf{f}_{\mathrm{ref}}$ from the generated shape and the reference, respectively, and compute their cosine similarity:
\begin{equation}
    \mathrm{ULIP\text{-}2~Sim.} = \frac{\mathbf{f}_{\mathrm{gen}} \cdot \mathbf{f}_{\mathrm{ref}}}{\|\mathbf{f}_{\mathrm{gen}}\| \;\|\mathbf{f}_{\mathrm{ref}}\|}.
\end{equation}
Higher ULIP-2 Similarity indicates that the generated shape is more semantically consistent with the reference.

\paragraph{Mean Curvature Sum (MCS).}
MCS quantifies the geometric richness and surface detail of a generated mesh without requiring a ground-truth reference. We first normalize the mesh to a unit bounding box centered at the origin and compute the mean curvature $H_i$ at each vertex $i$ via the cotangent Laplacian~\cite{meyer2003discrete}. The MCS is then computed as the area-weighted average of absolute mean curvatures over all vertices:
\begin{equation}
    \mathrm{MCS} = \frac{\sum_{i=1}^{N} A_i \,|H_i|}{\sum_{i=1}^{N} A_i},
\end{equation}
where $N$ is the number of vertices and $A_i$ is the barycentric area associated with vertex $i$. Higher MCS indicates richer surface detail and more geometric complexity.

\paragraph{Token and Time Efficiency.}
For all methods, we report the average runtime over 1,000 cases on an RTX 4090. For each method, we count the number of tokens required to generate the shape at the highest resolution or fidelity level. For example, for OctGPT, we use the number of tokens corresponding to the highest-resolution mesh from the final regression step.

\paragraph{Global Codebook Utilization (GCU).}
GCU measures the fraction of the entire codebook that is actively used across the full dataset. Given a codebook of size $C$ and $N$ tokenized shapes, let $\mathcal{U} = \bigcup_{n=1}^{N} \{s_1^{(n)}, \dots, s_{L_n}^{(n)}\}$ denote the set of all distinct codebook indices observed across all shapes. GCU is defined as:
\begin{equation}
    \mathrm{GCU} = \frac{|\mathcal{U}|}{C}.
\end{equation}
A higher GCU indicates more effective utilization of the codebook capacity, while a low GCU signals codebook collapse where large portions of the codebook remain unused.

\paragraph{Average Per-Shape Codebook Utilization (APCU).}
APCU captures how diversely each individual shape utilizes the codebook. For shape $n$ with token sequence $\mathbf{s}^{(n)} = (s_1, \dots, s_{L_n})$, let $\mathcal{U}^{(n)} = \{s_1^{(n)}, \dots, s_{L_n}^{(n)}\}$ denote the set of distinct codebook indices used by that shape. APCU is defined as the ratio of distinct indices to the sequence length, averaged over all shapes:
\begin{equation}
    \mathrm{APCU} = \frac{1}{N}\sum_{n=1}^{N} \frac{|\mathcal{U}^{(n)}|}{L_n}.
\end{equation}
A higher APCU indicates that each shape uses a richer variety of codebook entries, whereas a low APCU suggests that individual shapes are dominated by a small number of repeated codes.

\paragraph{Spatial Ordering Compressibility Gap ($\Delta_{\text{gzip}}$).}
To quantify whether spatial sorting induces learnable sequential structure in the token sequence, we measure the compressibility gain of sorted sequences over randomly permuted ones using gzip~\cite{ziv1977universal}, following recent work that employs general-purpose compressors as non-parametric entropy estimators~\cite{jiang2023low,deletang2024language}. For each shape $n$ with spatially sorted token sequence $\mathbf{s}^{(n)} = (s_1^{(n)}, \dots, s_{L_n}^{(n)})$, we compute the gzip compression rate (in bits per token):
\begin{equation}
    R^{(n)} = \frac{8 \cdot |\texttt{gzip}(\mathbf{s}^{(n)})|}{L_n},
\end{equation}
where $|\texttt{gzip}(\cdot)|$ denotes the compressed size in bytes. We similarly compute the average compression rate over $K$ random permutations $\pi_k$ of the same token \emph{set}:
\begin{equation}
    \bar{R}_{\text{rand}}^{(n)} = \frac{1}{K} \sum_{k=1}^{K} \frac{8 \cdot |\texttt{gzip}(\pi_k(\mathbf{s}^{(n)}))|}{L_n}.
\end{equation}
The Spatial Ordering Compressibility Gap is defined as:
\begin{equation}
    \Delta_{\text{gzip}} = \frac{1}{N} \sum_{n=1}^{N} \left(\bar{R}_{\text{rand}}^{(n)} - R^{(n)}\right).
\end{equation}
A larger $\Delta_{\text{gzip}}$ indicates that spatial sorting produces more compressible sequences than random orderings, implying the existence of exploitable sequential structure for autoregressive modeling. A $\Delta_{\text{gzip}} \approx 0$ suggests that the token sequence exhibits little spatial structure, indicating that autoregressive modeling may not benefit from the chosen ordering.

\clearpage
\section{Baseline Settings} \label{app:sec:baselines}

We evaluate all baselines using their official implementations and default configurations unless otherwise noted.

\paragraph{BrickGPT~\cite{pun2025generating}.}
BrickGPT fine-tunes Llama-3.2-1B-Instruct with LoRA ($r{=}32$, $\alpha{=}16$) to autoregressively generate LEGO-style brick assemblies at $20^3$ resolution from text prompts.
We cap both rejection sampling and physics-informed rollback at 50 iterations to balance generation quality and efficiency.

\paragraph{OctGPT~\cite{wei2025octgpt}.}
OctGPT adopts octree-based tokenization at $256^3$ resolution and generates shapes through next-scale prediction.
We employ the publicly released checkpoint pretrained on Objaverse for text-conditioned generation with an octree depth of 8.
The iteration counts per depth level (depths 3--6) are set to [64, 128, 128, 256], with starting temperatures of [1.0, 1.2, 0.5, 0.5], respectively.

\paragraph{CraftsMan3D~\cite{li2024craftsman}.}
We adopt the CraftsMan3D variant equipped with DoraVAE, which encodes shapes into 2048 latent tokens for improved geometric detail.
Meshes are extracted at $512^3$ resolution with 50 sampling steps and a classifier-free guidance (CFG) scale of 7.5.

\paragraph{TRELLIS~\cite{xiang2024structured}.}
We use the official \texttt{TRELLIS-image-large} checkpoint with its default configuration.
TRELLIS follows a two-stage flow matching pipeline: sparse structure generation (25 steps, CFG scale 7.5) followed by structured latent generation (25 steps, CFG scale 3.0), with meshes decoded from 3D Gaussian splatting representations.

\paragraph{Direct3D-S2~\cite{wu2025direct3d}.}
Direct3D-S2 employs sparse voxel SDF representations with a three-stage cascaded diffusion pipeline operating at up to $1024^3$ resolution.
We use 50, 30, and 15 DDIM sampling steps for the dense, $512^3$, and $1024^3$ stages, respectively, with a CFG scale of 7.5 for all stages.

\paragraph{TRELLIS2~\cite{xiang2025native}.}
We use the official \texttt{TRELLIS.2-4B} checkpoint with the single-stage 1024 pipeline.
TRELLIS2 represents shapes as O-Voxel sparse voxels with 16$\times$ spatial downsampling, first generating a sparse structure at $64^3$ latent resolution and then producing shape latents at $1024^3$.
All stages employ 50 Flow Euler sampling steps with a CFG scale of 7.5.

\paragraph{SuperVoxelGPT.}
We describe the data processing pipeline and architecture settings of SuperVoxelGPT.

\paragraph{Data Processing.}
We adopt the same rendering setup as TRELLIS2 for multi-view image generation and use the captions from Trellis-500K as multimodal prompts. Unlike TRELLIS2, whose first stage predicts only binary occupancy, our Stage~1 additionally predicts per-voxel \emph{saliency}---a measure of local geometric complexity. Specifically, we compute mesh saliency on the original surface and assign each occupied voxel the maximum saliency value among its interior mesh vertices. The resulting $64^3$ saliency volume is upsampled to $256^3$ via cubic interpolation, from which supervoxel centers are computed through GPU-accelerated Centroidal Voronoi Tessellation (CVT)~\cite{zheng2020computing}. For Stage~2, we follow the same output representation as TRELLIS2, using its O-Voxel algorithm to compute ground-truth shape labels and decode the generated structure into meshes.

\paragraph{Architecture.}
Our framework employs two VQ-VAE architectures. The \emph{Saliency VQ-VAE} compresses a $64^3$ saliency volume into $8^3 \times 2 = 1{,}024$ tokens via three downsampling stages and two residual quantizers. The \emph{SuperVoxelVAE} inherits the VAE backbone of TRELLIS2 but introduces plug-in Grid-to-Supervoxel (knn neighborhood size 16), FSQ (quantization levels [9, 9, 5, 5, 5]), and Supervoxel-to-Grid (knn neighborhood size 16) layers that encode shape geometry into one token per supervoxel. During training of both VQ-VAE models, we randomly overwrite 0--5\% of tokens with random indices to ensure robustness against noise. To maximize inference speed, we adopt MaskGIT for parallel prediction of all 1{,}024 saliency tokens in Stage~1, which is among the fastest generation algorithms available. Moreover, since saliency volume prediction provides the sequence length for Stage~2, we directly apply Jacobi decoding implementation~\cite{santilli2023accelerating} to generate the supervoxel token sequence in parallel.

\section{Parameter Sensitivity Analysis} \label{app:sec:sensitivity}

\begin{figure*}[t]
    \centering
    \begin{overpic}[width=\linewidth]{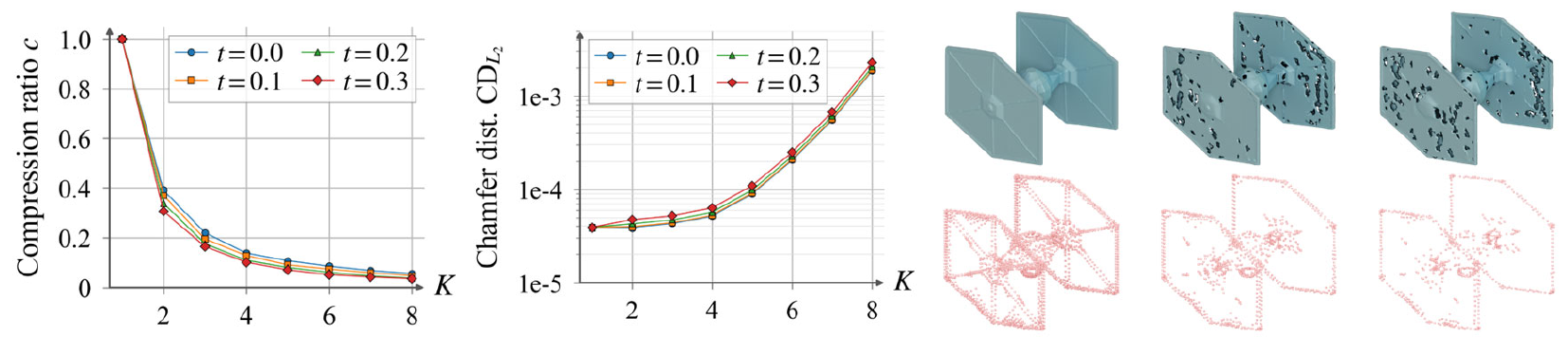}
        \put(13,-3){\small (a)}
        \put(42,-3){\small (b)}
        \put(65,-3){\small (c)}
        \put(80,-3){\small (d)}
        \put(93,-3){\small (e)}
    \end{overpic}
    \vspace{2mm}
    \caption{Parameter sensitivity analysis on 1,000 test shapes. (a)~Compression ratio $c$ and (b)~encode-decode CD$_{L_2}$ for varying $(K,t)$; lower is better. (c)--(e) show the generated shape (top) and the corresponding supervoxel centroids (bottom) at $(K{=}4,t{=}0.1)$, $(K{=}6,t{=}0.1)$, and $(K{=}8,t{=}0.1)$, respectively.}
    \label{fig:tokenizer_sensitivity}
\end{figure*}

As a token compression method, our saliency-based CVT relies on two important parameters to control the compression ratio. Specifically, $K\!\in\![1,64]$ controls the maximum cell size in low-saliency regions and therefore determines the upper bound of local compression, while $t\!\in\![0,1]$ is the saliency threshold used to suppress small saliency-estimation noise. In this section, we focus on how $K$ and $t$ affect both compression ratio and reconstruction quality. We sweep $K\!\in\!\{1,\dots,8\}$ and $t\!\in\!\{0,0.1,0.2,0.3\}$, retrain and evaluate 32 SuperVoxelVAE variants, and report compression and encode-decode reconstruction performance on 1,000 test shapes in \cref{fig:tokenizer_sensitivity}.

The results show a clear trade-off between compactness and reconstruction fidelity. Larger $K$ and $t$ produce fewer supervoxel cells and thus stronger compression, but they also increase reconstruction error because more fine-grained geometric regions are merged into coarse cells. The setting $K{=}4$ lies near the knee of the compression curve while preserving reconstruction quality: it benefits from the sweet-spot where non-uniform detail distribution yields a rapid reduction in token count, while avoiding the over-compression of salient regions observed with larger values such as $K{=}6$ or $K{=}8$. Meanwhile, $t{=}0.1$ provides additional compression with nearly unchanged reconstruction quality compared with $t{=}0$. We therefore use $(K{=}4,t{=}0.1)$ as the default setting in our experiments.

\begin{figure}[t]
    \centering
    \begin{overpic}[width=\linewidth]{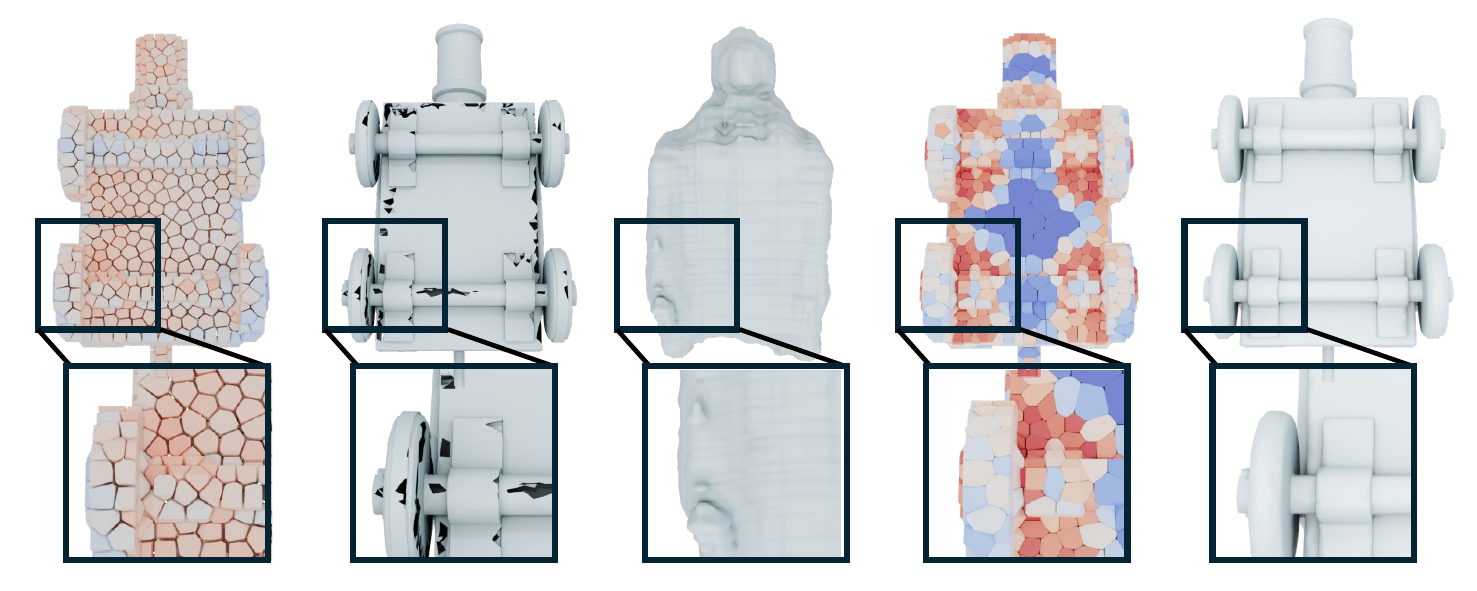}
        \put(8,-3){\small (a)}
        \put(28,-3){\small (b)}
        \put(48,-3){\small (c)}
        \put(68,-3){\small (d)}
        \put(88,-3){\small (e)}
    \end{overpic}
    \vspace{4mm}
    \caption{Qualitative ablation results. (a)~Uniform CVT supervoxel partition (Model~2). (b)~Generation result with uniform CVT. (c)~Generation result from CraftsMan3D plus FSQ layer (Model~3). (d)~Our saliency-guided CVT supervoxel partition. (e)~Our result. The color definition of saliency is the same as Fig.1 in the main paper.}
    \label{fig:ablation_vis}
\end{figure}

\clearpage
\section{Ablation Studies} \label{app:sec:ablation}
Our method introduces three key design choices: a two-stage MLLM framework, saliency-guided CVT for adaptive supervoxel construction, and a KNN-based grid-to-supervoxel conversion in SuperVoxelVAE. We ablate each component to validate its effectiveness. All ablation experiments are evaluated on the image-to-3D task using the same 1{,}000-shape test set. \cref{fig:ablation_vis} provides a qualitative comparison, and \cref{tab:ablation_jacobi,tab:ablation_cvt,tab:ablation_knn} report quantitative results.

\begin{table}[t]
    \centering
    \caption{Ablation on decoding strategy. Token-by-token decoding (Model~1) and Jacobi decoding (Ours) achieve comparable generation quality and token prediction accuracy, while Jacobi decoding is significantly faster. Token Acc.\ measures the percentage of predicted tokens that match the ground-truth token indices produced by SuperVoxelVAE encoding.}
    \label{tab:ablation_jacobi}
    \tablestyle{4pt}{1.1}
    \resizebox{\linewidth}{!}{
    \begin{tabular}{l|cccccc}
        \toprule
        Decoding Strategy & CD$_{L2}$ $\downarrow$ & PSNR $\uparrow$ & ULIP-2 Sim. $\uparrow$ & MCS $\uparrow$ & Token Acc.\ (\%) $\uparrow$ & Infer Time (s) $\downarrow$ \\
        \midrule
        Token-by-Token (Model 1)  & \textbf{0.0122} & \textbf{32.09} & \textbf{0.47} & \textbf{44.21} & \textbf{97.80} & 47.21 \\
        Jacobi Decoding (Ours)    & \textbf{0.0122} & 32.08 & \textbf{0.47} & \textbf{44.21} & 97.50 & \textbf{\phantom{0}1.20} \\
        \bottomrule
    \end{tabular}
    }
\end{table}

\begin{table}[t]
    \centering
    \caption{Ablation on CVT strategy at the same compression ratio. Saliency-guided CVT (Ours) significantly outperforms uniform CVT (Model~2) by concentrating tokens in geometrically complex regions.}
    \label{tab:ablation_cvt}
    \tablestyle{6pt}{1.1}
    \resizebox{0.75\linewidth}{!}{
    \begin{tabular}{l|cccc}
        \toprule
        CVT Strategy & CD$_{L2}$ $\downarrow$ & PSNR $\uparrow$ & ULIP-2 Sim. $\uparrow$ & MCS $\uparrow$ \\
        \midrule
        Uniform CVT (Model 2)       & 0.0157 & 29.09 & 0.469 & 40.13 \\
        Saliency CVT (Ours)         & \textbf{0.0122} & \textbf{32.08} & \textbf{0.474} & \textbf{44.21} \\
        \bottomrule
    \end{tabular}
    }
\end{table}

\begin{table}[t]
    \centering
    \caption{Ablation on grid-to-supervoxel encoding. We compare our spatially localized encoding method against CraftsMan3D~\cite{li2024craftsman} with FSQ quantization (Model~3). GCU, APCU, and $\Delta_{\text{gzip}}$ measure codebook utilization and spatial ordering compressibility (see \cref{app:sec:metrics}).}
    \label{tab:ablation_knn}
    \tablestyle{4pt}{1.1}
    \resizebox{\linewidth}{!}{
    \begin{tabular}{l|ccccccc}
        \toprule
        Encoding Strategy & CD$_{L2}$ $\downarrow$ & PSNR $\uparrow$ & ULIP-2 Sim. $\uparrow$ & MCS $\uparrow$ & GCU (\%) $\uparrow$ & APCU $\uparrow$ & $\Delta_{\text{gzip}}$ $\uparrow$ \\
        \midrule
        CraftsMan3D + FSQ (Model 3) & 0.0402 & 20.91 & 0.364 & 21.73 & 99.97 & 0.5649 & 0.0060 \\
        Ours                         & \textbf{0.0122} & \textbf{32.08} & \textbf{0.474} & \textbf{44.21} & \textbf{100.00} & \textbf{0.7727} & \textbf{0.1254} \\
        \bottomrule
    \end{tabular}
    }
\end{table}

\paragraph{Two-Stage Design: Jacobi Decoding vs.\ Token-by-Token Decoding.}
Following TRELLIS2~\cite{xiang2025native}, we adopt a two-stage MLLM framework. The primary motivation is that the first stage provides a coarse-grained prediction of the supervoxel structure, which determines the output sequence length and 3D structure for the second stage and thereby enables parallel inference via Jacobi decoding~\cite{santilli2023accelerating}.
To validate this design, we construct \textbf{Model~1} by removing Jacobi decoding and reverting to standard token-by-token autoregressive inference with KV cache. On the 1{,}000-shape test set, token-by-token decoding requires an average of 47.21\,s for the MLLM stage alone, whereas our Jacobi decoding completes in only 1.20\,s. As shown in \cref{tab:ablation_jacobi}, the two decoding strategies achieve nearly identical generation quality in the final reconstruction task. This is because we inject noise into the token indices when training the VQ-VAE, which ensures the decoder is robust to imperfect sequence predictions. As a result, Jacobi decoding achieves comparable performance to token-by-token decoding while greatly improving inference speed.

\paragraph{Saliency-Guided CVT vs.\ Uniform CVT.}
Saliency-guided CVT is a core design of our framework, determining \emph{where} to allocate more tokens for geometrically complex regions while avoiding token waste on smooth surfaces. To validate its importance, we construct \textbf{Model~2}: we keep the same compression ratio but run CVT on a \emph{uniform} saliency volume (\ie, setting all saliency values to a constant), producing uniformly distributed supervoxel centers (\cref{fig:ablation_vis}(a)). We then retrain both the SuperVoxelVAE and the MLLM on this uniform partition from scratch.
As shown in \cref{tab:ablation_cvt}, replacing saliency-guided CVT with uniform CVT leads to a notable drop in generation quality, particularly for shapes with highly non-uniform geometric complexity (\cref{fig:ablation_vis}(a,b) vs.\ (d,e)). This confirms our core insight: geometrically complex regions require a denser token allocation to faithfully preserve fine-grained structures, and saliency-guided partitioning is essential for achieving this under a fixed token budget.

\paragraph{SuperVoxel-Based Compression vs. Set-based Compression.}
SuperVoxelVAE compresses 3D information through a spatially localized pipeline: sparse convolutions first encode the occupancy grid into grid features, KNN Cross Attention~\cite{wu2022point} ($K{=}16$) then aggregates neighboring grid features into supervoxel features, and finally FSQ quantizes the supervoxel features into discrete tokens. Crucially, the entire process operates locally---each grid cell only interacts with its neighboring supervoxel centers rather than the entire shape---thereby preserving the correspondence between each token and its spatial position. In contrast, set-based methods such as CraftsMan3D~\cite{li2024craftsman} employ global cross-attention to compress shape information into a set of continuous, unordered latent features. This holistic encoding destroys the token--position correspondence, \ie, the sequential structure of tokens is lost, rendering such representations unsuitable for position-ordered autoregressive generation. To validate this analysis, we construct \textbf{Model~3} using the same MLLM architecture as ours, but replacing our SuperVoxelVAE with the set-based encoder-decoder from CraftsMan3D~\cite{li2024craftsman} and inserting an identical FSQ layer (levels $[9,9,5,5,5]$) at the bottleneck to quantize the latent features into discrete tokens. We then retrain both the VAE and the MLLM from scratch under the same training configuration.
As shown in \cref{tab:ablation_knn} and \cref{fig:ablation_vis}(c), our supervoxel-based encoding significantly outperforms the set-based alternative across all metrics, confirming that the preservation of spatial locality is essential for autoregressive 3D generation.

\section{Discussion} \label{app:sec:discussion}

We note that concurrent Gaussian-based methods~\cite{lan2025gaussiananything} have demonstrated strong performance in novel-view synthesis and compact 3D representation. We regard these methods as a parallel direction to our mesh-generation setting rather than direct substitutes. Gaussian representations are primarily optimized for differentiable rendering and view synthesis, and are not specifically designed for extracting clean, high-quality meshes with reliable topology. We therefore examine this issue under our mesh evaluation protocol. We also discuss the training scalability of our method.

\paragraph{Gaussian Splatting-based methods.}
\begin{wrapfigure}{r}{0.5\linewidth}
    \vspace{-12pt}
    \centering
    \begin{overpic}[width=\linewidth]{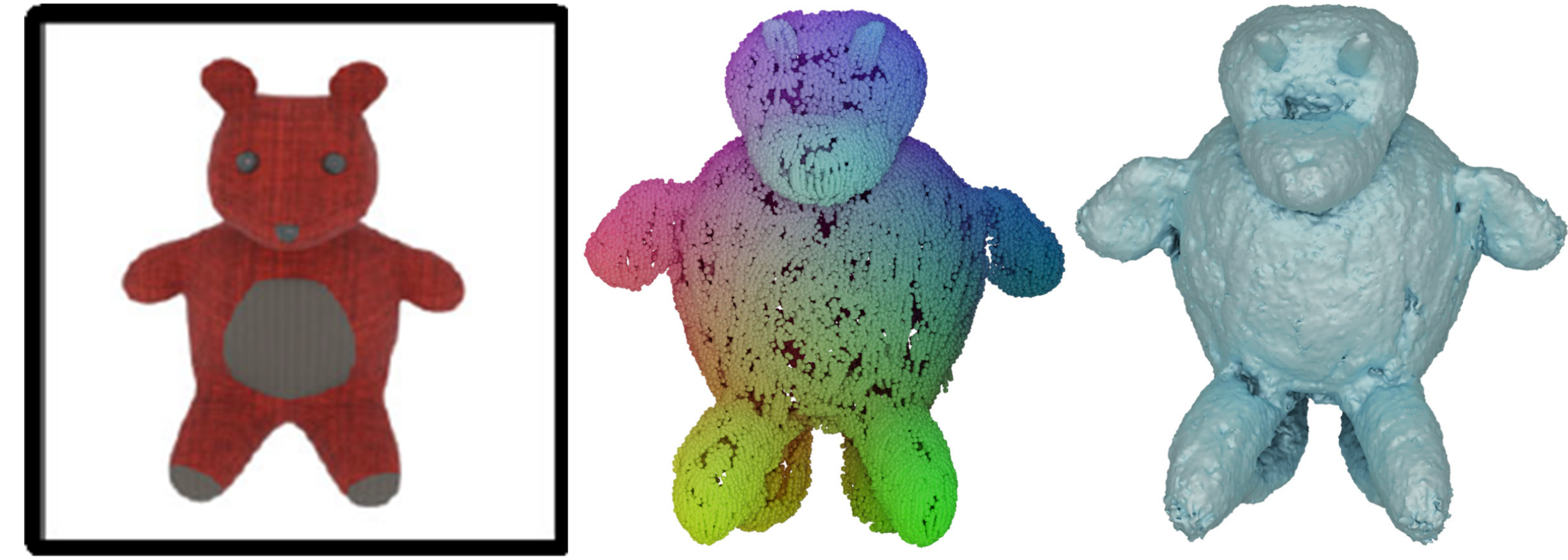}
        \put(15,-7){\small (a)}
        \put(48,-7){\small (b)}
        \put(81,-7){\small (c)}
    \end{overpic}
    \vspace{2mm}
    \caption{GaussianAnything outputs on an image-to-shape example. From left to right: (a)~the input image, (b)~the predicted Gaussian centers, and (c)~the extracted (TSDF-reconstructed) mesh.}
    \label{fig:app_gaussian_example}
    \vspace{-8pt}
\end{wrapfigure}
We additionally evaluate GaussianAnything~\cite{lan2025gaussiananything}, a Gaussian Splatting (GS)-based image-to-shape method, on our image-to-shape test set. Since our target output is an explicit mesh, we convert its predicted Gaussians to meshes through their TSDF reconstruction method before applying the same evaluation protocol. As shown in \cref{fig:app_gaussian_example}, the reconstructed meshes exhibit noisy surfaces and weak topology: GaussianAnything obtains CD$_{L_2}{=}0.0471$, PSNR${=}20.33$, ULIP-2 Sim.${=}0.363$, and an average runtime of $57.92$\,s, whereas our method obtains CD$_{L_2}{=}0.0122$, PSNR${=}32.08$, ULIP-2 Sim.${=}0.474$, and $4.60$\,s on the same test set. These results support our view that GS-based representations are complementary to mesh-generation methods: methods based on this representation can provide texture generation, which is beyond the scope of our method, but heavy surface noise and numerous holes remain a major challenge for them. In the future, exploring token-reduction methods that can support both meshes and textures is an important research direction.

\begin{table}[t]
    \centering
    \caption{Maximum training cost of representative methods. ``--'' means the training time is not reported by the corresponding method.}
    \label{tab:app_training_cost}
    \tablestyle{3pt}{1.1}
    \resizebox{\linewidth}{!}{
    \begin{tabular}{l|cccccc}
        \toprule
        Method & TRELLIS~\cite{xiang2024structured} & TRELLIS2~\cite{xiang2025native} & Direct3D-S2~\cite{wu2025direct3d} & OctGPT~\cite{wei2025octgpt} & BrickGPT~\cite{pun2025generating} & Ours \\
        \midrule
        GPU  & 64$\times$A100 & 48$\times$H100 & 8$\times$A100 & 8$\times$4090 & 8$\times$A6000 & \textbf{8$\times$H100} \\
        Time & $>$7--14\,d & -- & $>14$\,d & 7\,d & 12\,h & \textbf{5\,d} \\
        \bottomrule
    \end{tabular}
    }
\end{table}

\paragraph{Training scalability.}
As shown in \cref{tab:app_training_cost}, SuperVoxelGPT trains with a moderate computational budget among high-resolution 3D generation methods: it uses 8$\times$H100 GPUs for 5 days with our current settings. Although OctGPT and BrickGPT report lower training cost, they are restricted to specific domains, \ie, specified shape categories or LEGO structures, respectively. In contrast, SuperVoxelGPT targets general high-resolution mesh generation, and its supervoxel tokenizer substantially shortens the 3D token sequence that the generative model needs to learn. This compact tokenization reduces the training burden of the MLLM stage and makes scaling to larger MLLMs and larger datasets more practical. We regard such large-scale training as complementary to our core contribution and leave it as an important direction for future work.

\clearpage
\section{Additional Qualitative Results} \label{app:sec:qualitative}

\begin{figure*}[!htbp]
    \centering
    \begin{overpic}[width=\linewidth]{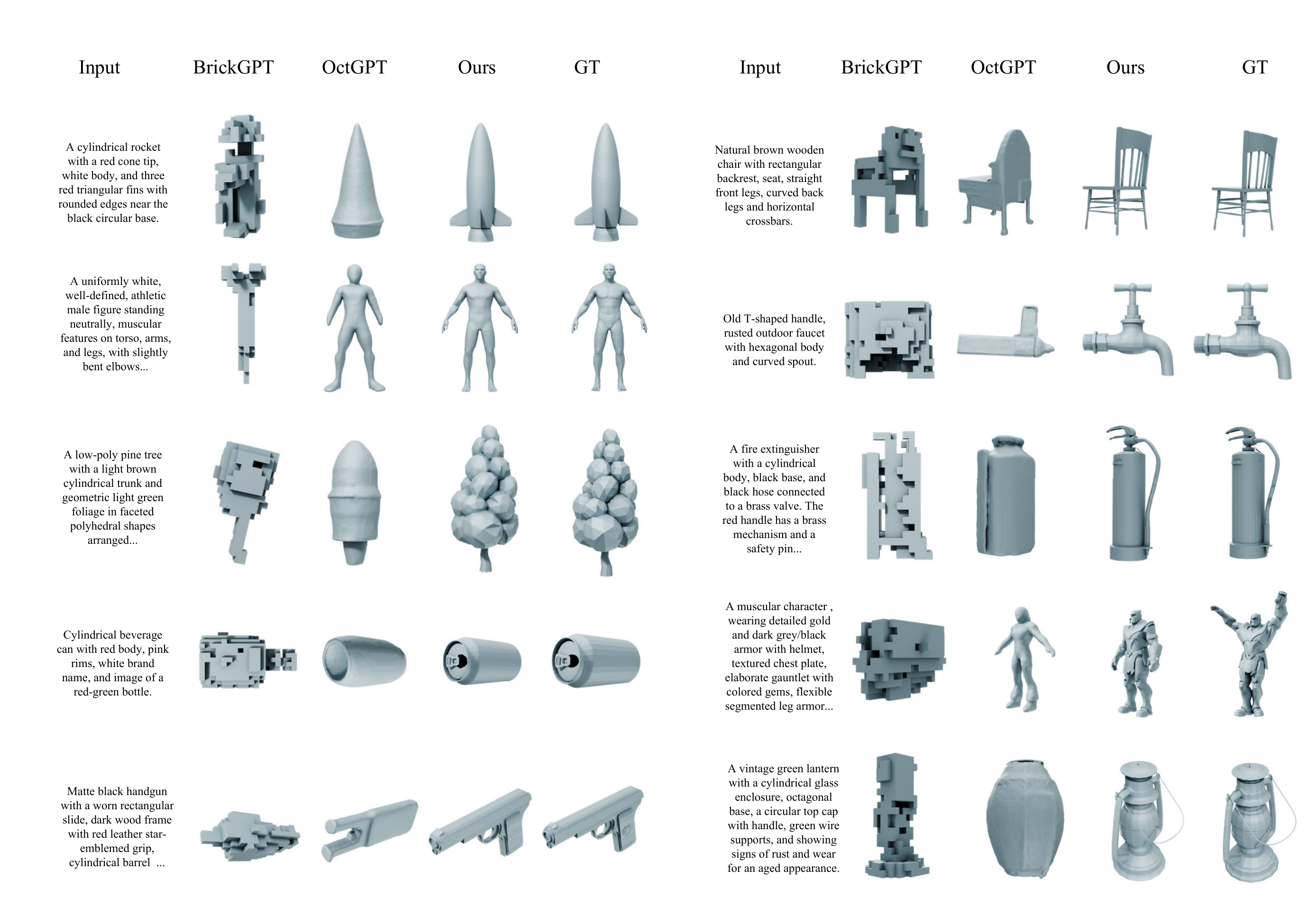}
        \put(0,55){\small (a)}
        \put(0,43.25){\small (b)}
        \put(0,31.5){\small (c)}
        \put(0,19.75){\small (d)}
        \put(0,8){\small (e)}
        \put(50,55){\small (f)}
        \put(50,43.25){\small (g)}
        \put(50,31.5){\small (h)}
        \put(50,19.75){\small (i)}
        \put(50,8){\small (j)}
    \end{overpic}
    \caption{Qualitative comparison on text-to-3D generation. (a)--(j) represent 10 different cases. Each case shows the input text, results from BrickGPT~\cite{pun2025generating} ($20^3$), OctGPT~\cite{wei2025octgpt} ($256^3$), our SuperVoxelGPT ($1024^3$), and the ground truth.}
    \label{fig:text2shape}
\end{figure*}

\cref{fig:text2shape} presents additional qualitative comparisons on the text-to-3D generation task. We compare our method against BrickGPT~\cite{pun2025generating} ($20^3$ resolution) and OctGPT~\cite{wei2025octgpt} ($256^3$ resolution) across 10 diverse text prompts with ground-truth (GT) references. BrickGPT produces coarse LEGO-style assemblies that lack geometric detail, while OctGPT generates smoother but still limited shapes. Our method generates significantly more detailed and geometrically faithful shapes at $1024^3$ resolution, closely matching the ground truth.

\begin{figure*}[t]
    \centering
    \includegraphics[width=\linewidth,height=0.95\textheight,keepaspectratio]{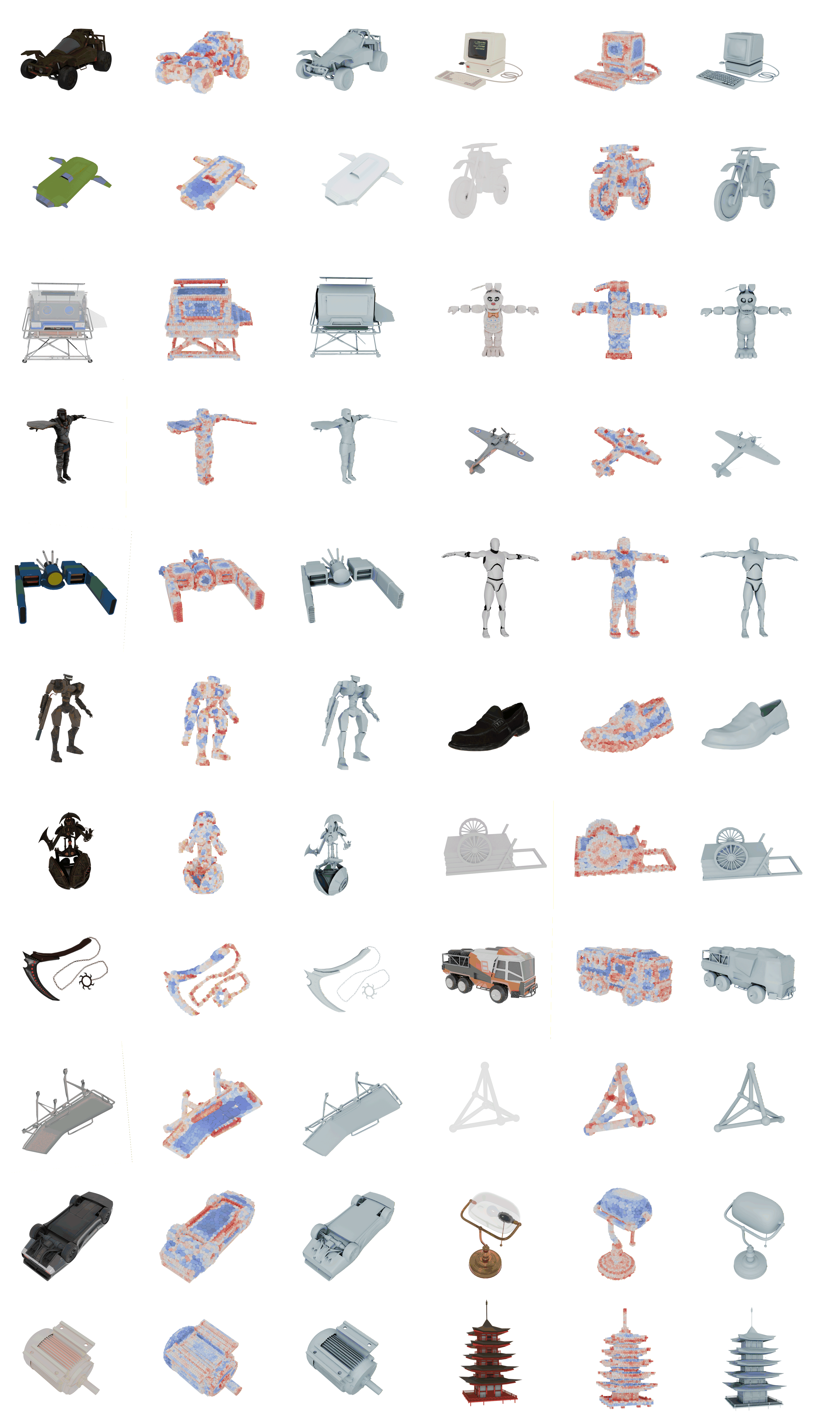}
\caption{Additional image-to-3D results. Each triplet shows, from left to right: the input image, the predicted supervoxel structure, and the decoded 3D mesh. The color definition of saliency is the same as Fig.1 in the main paper.}
    \label{fig:more_i2s}
\end{figure*}

\cref{fig:more_i2s} shows additional image-to-3D generation results from our method.

\end{document}